\documentclass[runningheads]{llncs}

\usepackage[T1]{fontenc}
\usepackage{graphicx}
\usepackage{amsmath,amssymb,amsfonts}
\usepackage{booktabs}
\usepackage{multirow}
\usepackage{siunitx}
\usepackage{xcolor}
\usepackage{textcomp}
\usepackage{subcaption}
\usepackage{enumitem}
\usepackage{xspace}
\usepackage[hidelinks]{hyperref}
\usepackage[nameinlink,capitalize]{cleveref}
\usepackage{silence}
\usepackage[nolist]{acronym}
\usepackage{tcolorbox}

\graphicspath{{./}}
\input{misc/commands_llncs}

\title{Adversarial Robustness of AI-Generated Image Detectors in the Real World}
\titlerunning{Adversarial Robustness of AI-Generated Image Detectors}

\author{Sina Mavali\inst{1} \and Jonas Ricker\inst{2} \and David Pape\inst{1} \and Asja Fischer\inst{2} \and Lea Sch{\"o}nherr\inst{1}}
\authorrunning{S. Mavali, J. Ricker, D. Pape, A. Fischer, and L. Sch{\"o}nherr}
\institute{
CISPA Helmholtz Center for Information Security, Saarbr{\"u}cken, Germany\\
\email{\{sina.mavali,david.pape,schoenherr\}@cispa.de}
\and
Ruhr University Bochum, Bochum, Germany\\
\email{\{jonas.ricker,asja.fischer\}@rub.de}
}

\begin{document}

\maketitle

\begin{abstract}
The rapid advancement of \ac{GenAI} is accompanied by a concerning rise in its misuse.
In particular, generated images are increasingly used to create visually convincing false narratives.
To counter this threat, a range of methods has been proposed to distinguish AI-generated images from authentic ones.
Yet their adversarial robustness is rarely assessed, and when it is, evaluations are often restricted to controlled rather than real-world settings.
In this work, we demonstrate that current state-of-the-art detectors are highly vulnerable to adversarial examples.
Most importantly, our experiments show that attacks remain effective even in realistic settings, where images are degraded throughout their lifecycle, \eg, common post-processing applied during uploads to social media platforms.
We also demonstrate the real-world impact of such vulnerabilities by successfully attacking a commercial detection tool.
Moreover, we investigate a defense based on robust CLIP features and show that it can improve robustness against adversarial perturbations.
While our results suggest that countermeasures can partially mitigate the risk of adversarial examples, we call upon model developers to make greater efforts toward the robust detection of AI-generated images.
The source code for this work is available at \url{https://github.com/dormant-neurons/aigi-break}.
\end{abstract}

\section{Introduction}

The creation of synthetic visual media is no longer a compelling vision, but has become part of everyday life and a vital business. 
In seconds, images with any desired content can be created using state-of-the-art generative models~\cite{midjourney_24,stable-diffusion,flux}. %
Recently, the quality of synthetic images has reached a level at which humans cannot distinguish them from real media~\cite{frank_2023}.
Thus, although \ac{GenAI} can support us in productive and creative tasks, the potential for misuse poses unprecedented challenges to our digital society, for example, by enabling the creation of fake identities on social media platforms~\cite{ricker-twitter-2024}.
In addition, AI-generated disinformation can discredit individuals, manipulate public opinion, and erode trust in democratic processes~\cite{ryan-mosleyHowGenerativeAI,drolsbach2025characterizingaigeneratedmisinformationsocial}. 
While model providers implement safety guardrails to prevent misuse, users often find loopholes.
For instance, journalists were able 
to create conspiracy imagery related to the 9/11 attacks and the assassination of John F.\ Kennedy using Google's Gemini~\cite{verge-nano-banana}.
Additionally, recent law-enforcement reports show that synthetic media is already being misused in many downstream abuse scenarios.
For instance, the FBI attributes nearly \$893 million in reported losses in 2025 to AI-related complaints and notes that scammers use fake social profiles, voice clones, forged identity documents, and realistic videos of public figures or loved ones~\cite{FBI2025IC3}.

Since \acp{AIGI} are often widely shared on various social media platforms before being removed~\cite{direstaHowSpammersScammers2024}, they pose a serious risk to global political discourse and public trust. Due to this growing threat, detecting \acp{AIGI} is an active research topic.
Proactive methods such as watermarking~\cite{fernandez-23-stable,fei-22-supervised} and cryptographically signed metadata in image files~\cite{c2pa} pose significant challenges: These methods require all parties involved to agree on the same standard and cooperate, while adversaries could use custom generative models or simply remove metadata or watermarks before distributing harmful images~\cite{jiang-23-evading,hu2025a,park2025chimera}. Another approach, which does not require the cooperation of model providers, is to passively detect synthetic images based on forensic artifacts. 
The majority of detectors utilize standard \acp{CNN}~\cite{cnnspot-20,grag-21} or pre-trained foundation models~\cite{univfd-23,drct-24} to distinguish between generated and real images. %
While these detectors often achieve very high accuracy, they are typically evaluated only in a \emph{clean} setting, without considering an attacker's attempts to bypass detection.
However, as previous work suggests, adversarial examples can dramatically reduce the performance of \ac{AIGI} detectors~\cite{carlini_2020,abdullah-24-analysis}.
Yet this prior work leaves several practical questions open. Recent work~\cite{eddoubi2025raiddatasettestingadversarial} focuses on constructing adversarial datasets, but does not analyze the quality-attack trade-off that determines whether perturbations remain stealthy. More critically, existing evaluations largely operate in idealized settings, ignoring the image degradations (e.g., compression, resizing) that occur when images are shared on social media~\cite{carlini_2020}. Furthermore, prior work has not validated these vulnerabilities against commercial detection systems. Finally, while adversarial defenses exist for other vision tasks, it remains unexplored whether they are useful for \ac{AIGI} detectors. %

In this work, we address these limitations and evaluate \ac{AIGI} detectors under conditions they face in the real world. In particular, we study whether adversarial examples remain effective after common social media platform degradations such as resizing, JPEG compression, and other post-processing.
Through extensive experiments with four detection methods, three \ac{AIGI} datasets, and five attack algorithms, we demonstrate that state-of-the-art detectors can be successfully deceived with minimal impact on visual quality and without any knowledge of the detector's internals.
Critically, most attacks remain effective if images are degraded, as typically happens when they are uploaded to a social media platform. %
In combination with the general performance loss caused by image post-processing, our results show that detectors are effectively unusable under realistic, adverse conditions.
To validate our findings, we confirm the effectiveness of our attacks on a popular proprietary \ac{AIGI} detector, HIVE.
While our results underline the vulnerability of current \ac{AIGI} detectors, we also demonstrate that their robustness can be significantly improved by leveraging robustly pre-trained feature extractors, a practical defense approach not previously explored for \ac{AIGI} detection.%

We summarize our findings as follows:
\begin{itemize}
    \item \textbf{State-of-the-art \ac{AIGI} detectors are vulnerable to adversarial examples.} For instance, the Diverse Input Attack can render detectors unusable, reducing the average AUROC from 87.5\% on clean inputs to approximately 0.0\%. Critically, the attacker does not need internal knowledge of the detector (\ie, a black-box setting).
    \item \textbf{Attacks remain effective even if images are altered in social media uploads} involving compression, resizing, etc. Under the strongest attack, detectors still collapse after high degradation, with average AUROC dropping to 28.2\%; degradation alone can already reduce clean average AUROC from 87.5\% to 66.6\%.
    \item \textbf{Adversarial examples are not only a theoretical risk, but can be used against real-world commercial detection tools} such as HIVE. In our case study, HIVE's average AUROC drops from 98.9\% on clean images to 76.6\% under attack; after responsible disclosure, the HIVE team confirmed the validity of our results.
    \item Despite these vulnerabilities, \textbf{we demonstrate that adversarial robustness can be achieved without training detectors from scratch.} By replacing standard feature extractors with robustly pre-trained backbones (RobustCLIP), AIGI detectors gain significant protection against strong attacks (\eg, AUROC increasing from 0.0 to 75.7 under the Diverse Input attack).%
\end{itemize}

\smallskip

Based on our findings, we urge researchers and model developers to consider adversarial robustness when evaluating their detectors and call for further research on the development of robust \ac{AIGI} misuse prevention.

\section{Related Work}

\paragraph{Detection of AI-Generated Images}
Due to the potentially harmful consequences of photorealistic synthetic images, various detection methods have been proposed. Among them, the most common approach for detecting generated images is to learn discriminative features directly from the data. Training \acp{CNN}, such as ResNet-50~\cite{resnet-15}, on real and generated images has been shown to be surprisingly effective~\cite{cnnspot-20,grag-21,corviDetectionSyntheticImages2023}. However, developing detectors that can adapt to unseen generative models is an ongoing challenge. Recently, several works~\cite{univfd-23,Cozzolino_2024_CVPR,drct-24} have shown that using large vision foundation models (e.g., CLIP~\cite{clip-21}) can help train such universal detectors. These methods train a relatively simple classifier on top of the features extracted by the pre-trained model, resulting in good generalization capabilities and higher robustness to perturbations. A possible explanation is that by using the pre-trained feature extractor, the detector does not overfit on specific artifacts of a certain generative model, but learns to identify more general properties of synthetic images~\cite{univfd-23}.  In our work, we focus on AIGI classifiers due to their generalizability and their inherent vulnerability to adversarial examples.
We refer to the recent work of Tariang~\etal~\cite{tariang2024syntheticimageverificationera} for a more detailed overview. 

\paragraph{Adversarial Robustness of AIGI Detectors} Pioneering work by Carlini \& Farid~\cite{carlini_2020} demonstrated that forensic classifiers were fundamentally vulnerable to both white-box and black-box attacks. Following this, a significant thread of research pursued customized attacks tailored specifically to the AIGI detection task. For instance, some methods generated high-quality adversarial images by optimizing within the latent manifold of a specific generator, such as StyleGAN~\cite{li_2021} or by adversarially modifying image attributes (\eg, exposure, noise, and blur) to minimize statistical differences between real and fake images~\cite{hou_2023}. Other approaches focused on removing generative artifacts through trace removal or retouching methods~\cite{liu_2023}. Abdullah~\etal~\cite{abdullah-24-analysis} propose a semantic-based attack that leverages a fine-tuned foundation model to craft adversarial samples by manipulating image content. 

All of these specialized attacks often suffer from critical limitations: Many are tightly coupled to specific GANs, rendering them less relevant for SOTA diffusion-based detectors, while others rely on impractical assumptions, such as requiring an attacker to possess deep knowledge of the image generation pipeline. We argue that the pursuit of novel attacks, while technically interesting, has obscured a more fundamental vulnerability. Our findings demonstrate that simpler, classical adversarial methods are not only sufficient but are often more effective, as shown in our attack comparison in \Cref{sec:results}. %
There have also been efforts to create adversarial datasets for AI-generated images. More recently, Eddoubi \etal~\cite{eddoubi2025raiddatasettestingadversarial} proposed a large dataset of adversarial examples for AIGI detectors. However, their work has two main limitations. First, the adversarial examples are generated with a large noise budget, resulting in poor perceptual quality that compromises their stealthiness. Furthermore, no trade-off between attack success and quality is explored. Second, it does not investigate whether these examples remain effective after realistic post-processing such as compression and resizing. %
 Our work addresses these limitations by evaluating attacks under realistic degradations and validating against a commercial system.

\section{Threat Model}
\label{sec:threat_model}

The general objective of the attacker is to create an adversarial example that bypasses \ac{AIGI} detection tools, for example, in a social media platform's moderation system or an individual user's verification tool.
For this, we consider a realistic black-box scenario where the attacker has no access to the target model's gradients or parameters.
Unlike white-box settings that assume full transparency, this threat model mirrors the knowledge gap an adversary faces when targeting proprietary social media detectors or commercial APIs.
In this setting, the attacker must rely on surrogate models to craft adversarial examples, exploiting the transferability of adversarial perturbations. This approach tests the generalization of attacks across different architectures (e.g., transferring from a CNN-based surrogate to a transformer-based target) and training data.

Furthermore, our threat model accounts for the constraints that real-world deployment might impose on an attacker.
Images uploaded to the internet inevitably undergo post-processing steps such as compression and resizing. Additionally, an attacker may deliberately degrade an image before uploading it (e.g., by blurring or adding noise) to hide forensic artifacts and increase the success chance of the attack.
Consequently, we evaluate attack success not just on adversarial examples, but under various levels of degradation.

Finally, we anticipate an adaptive defender who may employ defense mechanisms to harden their detectors. 
By evaluating attacks against both standard and robustified models, we provide a comprehensive view of the current security landscape for \ac{AIGI} detection.

\section{Methodology}
In this work, we systematically evaluate the performance of state-of-the-art detection methods under realistic attack scenarios. To this end, we have curated a diverse collection of datasets that includes images generated by various generative models alongside a wide range of real images. We further utilize a series of attacks and degradations to assess the effectiveness of current detection methods.

\subsection{Datasets}
Our analysis employs three different datasets: Synthbuster~\cite{synthbuster-2023}, Chameleon~\cite{yan2025sanitycheckaigeneratedimage}, and images generated by GPT-4o~\cite{yan2025gptimgevalcomprehensivebenchmarkdiagnosing}. These datasets collectively offer a broad spectrum of synthetic and real images, enabling a comprehensive evaluation of detection methods. \emph{Synthbuster} provides a controlled evaluation environment by pairing 1,000 uncompressed real photographs from RAISE-1k~\cite{raise-15} with synthetic counterparts generated by nine different diffusion models. This design mitigates dataset biases and allows for a granular analysis of detector performance across various generative architectures. To test detectors in a more challenging setting, we use the \emph{Chameleon} dataset. It contains high-resolution synthetic images that have passed a human perception ``Turing Test,'' making them perceptually indistinguishable from their real counterparts, which are sourced from online photo repositories. 
As the \emph{GPT-4o} dataset lacks a paired set of real images, we employ the real images from the Synthbuster dataset (RAISE-1k) for evaluation.

\subsection{Detectors}
For our evaluation, we consider four state-of-the-art \ac{AIGI} image detectors that have demonstrated high accuracy and robust generalization in detecting \ac{DM}-generated images: Corvi~\cite{grag-21}, UnivFD~\cite{univfd-23}, and DRCT~\cite{drct-24} (both Conv-B and CLIP versions). These detectors were chosen to span a diverse range of concepts, architectures, and training paradigms.  %

\paragraph{Corvi}
This detector leverages a ResNet-50~\cite{resnet-15} and relies on heavy data augmentation during training. The notable improvement is the removal of the initial downsampling layer, which preserves high-frequency information at the cost of a higher number of trainable parameters.

\paragraph{UnivFD}
Ojha~\etal~\cite{univfd-23} address universal fake image detection by leveraging a large vision foundation model, specifically CLIP-ViT-L/14~\cite{dosovitskiyImageWorth16x162020}, as a feature extractor. A single linear layer is then trained on top of these pre-extracted features, while the CLIP weights remain frozen. This decoupled training approach improves the detector's generalization.

\paragraph{DRCT}
Diffusion reconstruction contrastive training~\cite{drct-24} is a training paradigm that enhances existing \ac{AIGI} image detectors by applying a contrastive loss to difficult image pairs. These pairs consist of a real image and its reconstruction, created by passing it through a diffusion model's forward-and-backward process. This method, which enhances generalization, is supported by both ConvNeXt and CLIP backbones.

For DRCT-CLIP and DRCT-ConvB we use publicly available checkpoints trained on the SD-1.4 train subset from GenImage~\cite{genimage-23}. Since no matching checkpoints are available for UnivFD and Corvi, we retrain both models on the same dataset, following the authors’ instructions.

\subsection{Attacks}

The objective of an attack against AIGI detection is to generate a perturbed image~\(x_{\text{adv}}\) that remains nearly indistinguishable from the original generated image~\(x\) yet is classified by the detector as real if it is AI-generated and vice versa. The perturbed image~\(x_{\text{adv}}\) is called an adversarial example. The task of finding such adversarial examples can be formulated as a constrained optimization problem:
\[
x_{\text{adv}} = \underset{x'}{\arg\max} \, L(f(x'), y) \quad \text{s.t.} \quad \| x' - x \|_p \leq \epsilon,
\]
where \(y\) represents the ground-truth label, \(\| \cdot \|_p\) denotes the \(\ell_p\)-norm, and \(L(\cdot)\) is typically the cross-entropy loss. The parameter $\epsilon$ defines the maximum allowed distortion between~\(x\) and~\(x_{\text{adv}}\). Attacks can generally be grouped into white-box and black-box methods. Under the white-box scenario, the attacker has complete knowledge of the detector's parameters and architecture, allowing the direct use of gradient-based techniques. In contrast, black-box attacks restrict the adversary to using a surrogate model to approximate its gradients or querying the target model.

We evaluate our model under a realistic black-box threat model, generating adversarial examples with attacks known for high transferability. These are grouped into two categories: transfer-based and query-based attacks. Transfer-based attacks use a surrogate model to craft adversarial examples offline, which is more efficient and stealthy. The success of these attacks depends on the similarity between the surrogate(s) and the target models. In contrast, query-based attacks interact directly with the target model, using its output to guide the attack. We use five representative methods, split into four transfer-based and one query-based attack:
\begin{itemize}
    \item \textbf{Projected Gradient Descent (PGD)}~\cite{pgd-19}: An iterative attack that makes small, imperceptible changes to an image by following the gradient of the loss function. We compute adversarial examples on a surrogate model and use their transferability to attack the target model. %
    \item \textbf{Ensemble-based}~\cite{tramer2020ensemble}: Extends PGD by averaging the gradients from multiple surrogate models to improve transferability.
    \item \textbf{Diverse Input Attack}~\cite{xie2019improvingtransferabilityadversarialexamples}: Enhances the ensemble attack by applying random transformations, like resizing, to the input at each iteration.
    \item \textbf{Universal Perturbations}~\cite{moosavidezfooli2017universaladversarialperturbations}: Finds a single, image-agnostic perturbation designed to fool the detector on a wide range of inputs, which we use in an ensemble mode.
    \item \textbf{Query-Efficient Attack}~\cite{ilyas2018queryefficientblackboxadversarialexamples}: Estimates the gradient direction by sending a few queries with structured noise to the target model and observing the outcomes.
\end{itemize}

\subsection{Degradations}
To complement the adversarial threat model, we also study quality degradations that naturally arise on social media platforms. Xu \etal~\cite{xuProFakeDetectingDeepfakes2024} showed that the degradation induced by real-world social media platforms (\eg, Facebook, Twitter) can be simulated using a combination of common transformations. While testing on live platforms would be the ideal case, it is generally not feasible due to Terms of Service violations regarding API limits and the upload of AI-generated content. Furthermore, such testing lacks scientific reproducibility due to the possible changes in platform compression. Therefore, we rely on simulated degradation protocols as seen in previous research~\cite{cnnspot-20,sm_2018}. To assess detector performance under both clean and adversarial conditions in real-world scenarios, we adopt the degradation simulation introduced by Xu~\etal, which corresponds to typical and severe processing scenarios:
\begin{itemize}
\item \textbf{Medium Degradation}: Corresponds to the common degradation levels observed on social media platforms. Images are downsampled to 75\% of their original size and JPEG compression is applied with a quality factor of 50.
\item \textbf{High Degradation}: Represents severe image quality loss, simulating scenarios with aggressive compression or multiple retransmissions. We achieve this by combining four types of degradation: Gaussian blurring (kernel size $5 \times 5$, $\mu=1$, $\sigma=1$), downsampling to 50\% of the original size, heavy JPEG compression with a quality factor of 20, and the addition of Gaussian noise ($\mu=1$, $\sigma=0.1$).
\end{itemize}

\paragraph{Terminology} To clarify our experimental conditions, we define four image categories along two axes: adversarial perturbation and post-processing degradation. \textbf{Clean} images are unperturbed and without degradation. \textbf{Degraded} images are unperturbed but have undergone post-processing such as compression or resizing. \textbf{Adversarial} images contain adversarial perturbations but no degradation. Finally, \textbf{degraded adversarial} images are adversarial examples that have also been degraded, representing the realistic scenario where an attacker uploads a perturbed image to social media.

\subsection{Metrics}
\label{sec:metrics}
To evaluate detector performance and attack effectiveness, we use three complementary metrics. \textbf{AUROC} (Area Under the ROC Curve) serves as our primary metric since it is threshold-independent and enables fair comparison across detectors. For practical interpretability, we also report \textbf{Accuracy} (using a fixed decision threshold of 0.5) and \textbf{TPR@5\%FPR} (true positive rate when false positive rate is fixed at 5\%), representing an operating point where false alarms are minimized. We exclude Attack Success Rate (ASR) as it cannot distinguish between a powerful attack and a detector with weaker clean performance. Lower AUROC, Accuracy, and TPR@5\%FPR indicate weaker detector performance; for attacked samples, this corresponds to a more successful attack. %

To quantify the impact of attacks relative to clean performance, we report the \textbf{AUROC Drop Ratio}, defined as:
\begin{equation}
\text{AUROC Drop Ratio} = \frac{\text{AUROC}_{\text{clean}} - \text{AUROC}_{\text{attacked}}}{\text{AUROC}_{\text{clean}}}
\end{equation}
This metric normalizes the reduction in performance by the detector's initial performance, which allows a fair comparison across different detectors and degradation levels with varying initial performance. We report absolute AUROC in tables to show actual detection performance, while using AUROC Drop Ratio in visualizations to highlight relative attack effectiveness.

\section{Robustness Analysis}
\label{sec:results}

In this section, we present a comprehensive analysis of the adversarial robustness of four state-of-the-art detectors.
We perform a robustness analysis comprising five attacks in a black-box setting and investigate the
effectiveness and quality of the resulting adversarial examples. 
Next, we examine the impact of real-world image degradations (\eg, social media compression) on attack performance, followed by a case study targeting a closed-source online detection tool.
Finally, we demonstrate that deploying a simple defense mechanism can significantly enhance the robustness of the detectors against the strongest attack.

\subsection{Baseline Attack Evaluation}
We first examine the adversarial robustness of our AI-generated image detectors by testing their performance on adversarial examples
created using five attacks in the black-box setting.
For this, we start with evaluating adversarial examples that have not been subjected to any degradation.

\paragraph{Setup} 
We use the $\ell_\infty$ norm with $\epsilon = 8/255$ for all attacks and a step size of $2/255$ for all transfer-based attacks.
We run \ac{PGD}, ensemble-based, and diverse input attack for 10 steps, while the universal perturbation attack requires 50
steps due to the harder optimization problem. We validated this choice by comparing attack effectiveness across 5, 10, and 20 steps on the Synthbuster dataset (see \Cref{tab:step_convergence} in the Appendix). The results show negligible differences in AUROC (less than 1 percentage point), indicating that 10 steps is sufficient for convergence. 
For the query-efficient attack, we use a maximum of 500 queries and 50 samples per draw for the gradient estimation, alongside a noise scale $\sigma=0.005$ and a minimum step size of $\epsilon \cdot 0.01$.
We evaluate attacks on a balanced test set of 1,500 images: 500 images per dataset (Synthbuster, Chameleon, GPT-4o), with each subset containing 250 real and 250 AI-generated images. Adversarial perturbations are applied to both classes to flip the detector's prediction (i.e., making AI-generated images appear real and vice versa).
For Synthbuster, we limit our analysis to images generated by \ac{SD-1.4}, as detectors achieve high performance on this subset in the clean setting.

\begin{figure}[t]
    \centering
    \includegraphics[width=0.75\columnwidth]{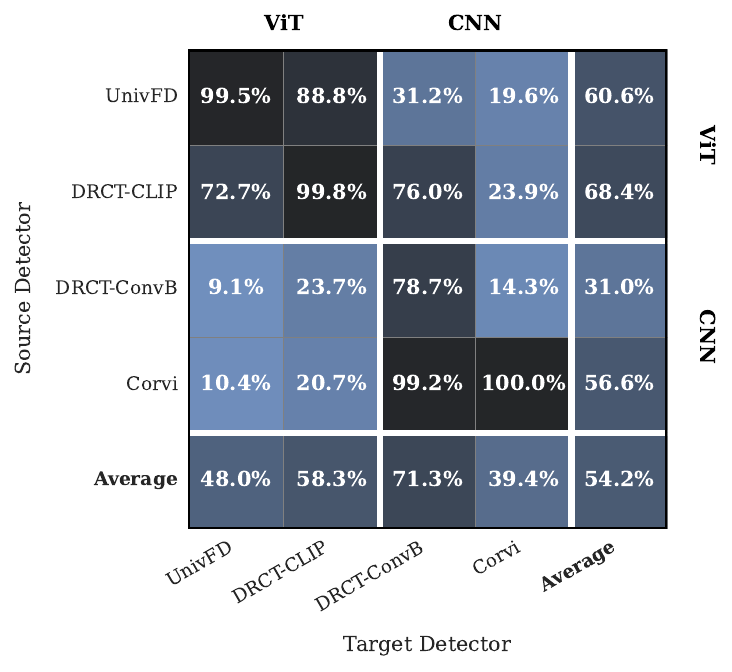}
    \caption{\textbf{Transferability Analysis (PGD, $\epsilon$=8/255).} Each cell shows the \textbf{AUROC Drop Ratio (\%)}. Darker colors indicate more effective attacks. Detectors are grouped by architecture: ViT-based (UnivFD, DRCT-CLIP) and CNN-based (DRCT-ConvB, Corvi). Diagonal cells show white-box attacks; off-diagonal cells show black-box transferability. The \textbf{Average} row/column summarizes overall effectiveness per detector.
}
    \label{fig:transfer_heatmap}
\end{figure}

\begin{table}[!htbp]
    \footnotesize
    \caption{\textbf{Detectors' Robustness.} AUROC, Accuracy, and TPR@5\%FPR averaged across all datasets for clean (no attack) and different attacks with varying degradation levels (None, Medium: 75\% downsampling and JPEG compression at quality 50, and High: Gaussian blur ($5\times5$, $\mu=1$, $\sigma=1$), 50\% downsampling, JPEG compression at quality 20, and Gaussian noise ($\mu=1$, $\sigma=0.1$)). ``LTO'' means Leave-Target-Out: attacks are optimized on all detectors except the evaluated target.}
    \setlength{\tabcolsep}{2pt}
    \newcommand{\mc}[1]{\multicolumn{3}{c}{#1}}
    \resizebox{\columnwidth}{!}{
    \begin{tabular}{@{\hspace{2pt}}lll|ccc|ccc|ccc@{\hspace{2pt}}}
        \toprule
        \multirow{2}{*}{\textbf{Attack}} & \multirow{2}{*}{\textbf{Target}} & \multirow{2}{*}{\textbf{Source}} & \mc{\textbf{AUROC}} & \mc{\textbf{Accuracy}} & \mc{\textbf{TPR@5\%FPR}} \\
        \cmidrule(lr){4-6} \cmidrule(lr){7-9} \cmidrule(lr){10-12}
        &&& \textbf{None} & \textbf{Med} & \textbf{High} & \textbf{None} & \textbf{Med} & \textbf{High} & \textbf{None} & \textbf{Med} & \textbf{High} \\
        \midrule
        \multirow{5}{*}{\textbf{Clean}}
        & UnivFD & {\textendash} & 91.4 & 89.5 & 89.7 & 80.9 & 79.2 & 78.5 & 65.9 & 63.1 & 62.4 \\
        & DRCT-CLIP & {\textendash} & 87.7 & 73.2 & 73.8 & 76.5 & 63.8 & 64.2 & 61.1 & 38.9 & 39.5 \\
        & DRCT-ConvB & {\textendash} & 85.6 & 80.4 & 46.8 & 75.1 & 70.3 & 52.1 & 51.6 & 43.2 & 12.8 \\
        & Corvi & {\textendash} & 85.4 & 77.4 & 55.9 & 70.7 & 64.3 & 53.8 & 60.1 & 48.7 & 24.3 \\
        \cmidrule{2-12}
        & Average & {\textendash} & 87.5 & 80.1 & 66.6 & 75.8 & 69.4 & 62.2 & 59.7 & 48.5 & 34.8 \\
        \midrule
        \midrule
        \multirow{13}{*}{\textbf{PGD}}
        & \multirow{3}{*}{UnivFD}
        & DRCT-CLIP & 18.0 & 61.5 & 64.0 & 29.8 & 58.1 & 55.1 & 0.1 & 16.7 & 13.5 \\
        && DRCT-ConvB & 80.5 & 85.1 & 68.3 & 71.5 & 72.6 & 57.9 & 39.2 & 50.5 & 20.9 \\
        && Corvi & 77.5 & 81.8 & 70.3 & 66.2 & 73.5 & 59.1 & 36.1 & 44.9 & 24.7 \\
        \cmidrule{2-12}
        & \multirow{3}{*}{DRCT-CLIP}
        & UnivFD & 7.5 & 38.5 & 59.1 & 30.4 & 49.0 & 50.6 & 0.0 & 4.5 & 11.7 \\
        && DRCT-ConvB & 62.3 & 68.5 & 62.5 & 52.8 & 50.4 & 51.3 & 14.7 & 21.2 & 14.1 \\
        && Corvi & 63.6 & 71.7 & 63.4 & 53.0 & 50.6 & 51.3 & 18.8 & 26.7 & 14.1 \\
        \cmidrule{2-12}
        & \multirow{3}{*}{DRCT-ConvB}
        & UnivFD & 58.5 & 53.6 & 47.0 & 50.5 & 49.9 & 48.1 & 14.0 & 9.3 & 4.7 \\
        && DRCT-CLIP & 16.4 & 45.1 & 46.9 & 44.5 & 47.7 & 48.6 & 0.8 & 4.5 & 3.5 \\
        && Corvi & 0.7 & 5.5 & 42.8 & 7.6 & 14.4 & 47.0 & 0.0 & 0.0 & 3.5 \\
        \cmidrule{2-12}
        & \multirow{3}{*}{Corvi}
        & UnivFD & 63.3 & 54.0 & 49.2 & 58.2 & 50.2 & 50.0 & 29.9 & 16.3 & 6.5 \\
        && DRCT-CLIP & 60.9 & 57.9 & 50.9 & 56.9 & 50.1 & 50.0 & 27.2 & 16.4 & 8.1 \\
        && DRCT-ConvB & 76.4 & 67.5 & 53.6 & 60.3 & 50.0 & 50.0 & 38.0 & 24.1 & 8.3 \\
        \cmidrule{2-12}
        & Average & {\textendash} & 48.8 & 57.6 & 56.5 & 48.5 & 51.4 & 51.6 & 18.2 & 19.6 & 11.1 \\
        \midrule
        \multirow{5}{*}{\textbf{Ensemble}}
        & UnivFD & LTO & 0.0 & 51.5 & 63.7 & 0.6 & 52.5 & 54.7 & 0.0 & 8.1 & 14.3 \\
        & DRCT-CLIP & LTO & 0.0 & 8.5 & 52.5 & 0.0 & 45.9 & 49.1 & 0.0 & 0.0 & 9.6 \\
        & DRCT-ConvB & LTO & 0.5 & 4.2 & 41.1 & 3.7 & 10.5 & 43.8 & 0.0 & 0.0 & 2.5 \\
        & Corvi & LTO & 0.0 & 1.2 & 42.7 & 0.1 & 41.5 & 50.0 & 0.0 & 0.0 & 5.6 \\
        \cmidrule{2-12}
        & Average & {\textendash} & 0.1 & 16.4 & 50.0 & 1.1 & 37.6 & 49.4 & 0.0 & 2.0 & 8.0 \\
        \midrule
        \multirow{5}{*}{\textbf{Diverse}}
        & UnivFD & LTO & 0.1 & 6.0 & 47.9 & 3.1 & 19.4 & 50.3 & 0.0 & 0.0 & 6.1 \\
        & DRCT-CLIP & LTO & 0.0 & 0.0 & 29.5 & 0.0 & 12.5 & 44.6 & 0.0 & 0.0 & 1.3 \\
        & DRCT-ConvB & LTO & 0.0 & 0.1 & 12.8 & 1.3 & 0.2 & 21.4 & 0.0 & 0.0 & 0.0 \\
        & Corvi & LTO & 0.0 & 0.0 & 22.5 & 0.1 & 8.6 & 49.8 & 0.0 & 0.0 & 0.8 \\
        \cmidrule{2-12}
        & Average & {\textendash} & 0.0 & 1.5 & 28.2 & 1.1 & 10.2 & 41.5 & 0.0 & 0.0 & 2.1 \\
        \midrule
        \multirow{5}{*}{\textbf{Universal}}
        & UnivFD & LTO & 28.6 & 79.4 & 65.3 & 35.6 & 69.1 & 57.6 & 0.8 & 34.8 & 16.3 \\
        & DRCT-CLIP & LTO & 0.0 & 62.1 & 60.9 & 0.5 & 50.1 & 50.5 & 0.0 & 12.5 & 14.4 \\
        & DRCT-ConvB & LTO & 0.1 & 12.5 & 41.7 & 2.4 & 24.6 & 43.8 & 0.0 & 0.0 & 2.3 \\
        & Corvi & LTO & 1.5 & 9.9 & 43.3 & 12.1 & 45.2 & 50.1 & 0.0 & 0.0 & 4.9 \\
        \cmidrule{2-12}
        & Average & {\textendash} & 7.6 & 41.0 & 52.8 & 12.7 & 47.2 & 50.5 & 0.2 & 11.8 & 9.5 \\
        \midrule
        \multirow{5}{*}{\textbf{Query}}
        & UnivFD & {\textendash} & 66.2 & 76.6 & 68.7 & 59.1 & 67.6 & 57.2 & 21.3 & 36.8 & 18.1 \\
        & DRCT-CLIP & {\textendash} & 71.1 & 68.6 & 62.8 & 55.1 & 50.7 & 51.4 & 28.0 & 23.2 & 13.9 \\
        & DRCT-ConvB & {\textendash} & 71.3 & 58.2 & 48.9 & 59.1 & 52.6 & 48.7 & 24.5 & 10.1 & 4.9 \\
        & Corvi & {\textendash} & 86.4 & 59.4 & 47.9 & 64.5 & 50.1 & 50.0 & 53.6 & 17.3 & 6.7 \\
        \cmidrule{2-12}
        & Average & {\textendash} & 73.8 & 65.7 & 57.1 & 59.5 & 55.2 & 51.8 & 31.9 & 21.9 & 10.9 \\
        \bottomrule
    \end{tabular}
    }
    \label{tab:attack_results_averaged}
\end{table}

\paragraph{Attack Comparison}
Table~\ref{tab:attack_results_averaged} presents our robustness analysis averaged across all datasets, reporting AUROC alongside Accuracy and TPR@5\%FPR. We evaluate the attacks and their transferability across different degradation levels, where ``None'' refers to attacks without degradation. ``LTO'' refers to ``Leave-Target-Out'': For each target detector, the ensemble attack is optimized on all other detectors and then evaluated on the held-out target.  Standard \textbf{\ac{PGD}} attacks, when transferred between models, are largely ineffective, though exceptions occur when the source and target models are well aligned (e.g., \ac{PGD} attacks from DRCT-CLIP significantly impact UnivFD, dropping the AUROC from 91.4 to 18.0). However, attacks from a different source, such as DRCT-ConvB, are less effective (AUROC drops to 80.5). We analyze \ac{PGD} transferability in \Cref{fig:transfer_heatmap}, highlighting strong transfers between CLIP-based detectors (UnivFD and DRCT-CLIP) and weaker transfers from CNN-based sources (DRCT-ConvB, Corvi) to CLIP-based targets. While Corvi shows considerable robustness to transferred \ac{PGD} attacks, white-box analyses (diagonal of \Cref{fig:transfer_heatmap}) confirm that this does not imply inherent immunity when \ac{PGD} is directly optimized on the target. For a comprehensive breakdown of these results across each dataset, we refer to Table~\ref{tab:attack_results} in the Appendix.

To counter the limited transferability of individual \ac{PGD}, \textbf{Ensemble-based Attacks},
specifically a Leave-Target-Out (LTO) ensemble, prove more effective. Adversarial examples crafted against an ensemble of all detectors except the target
reduce the average AUROC across detectors and datasets to 14.0 (down from 48.8 for \ac{PGD}), 
though Corvi remains more resilient (AUROC 50.3). The \textbf{Diverse Input Attack}, an enhancement of this ensemble approach,
is even more successful, reducing the average AUROC to a mere 2.0 and significantly degrading Corvi's performance 
(AUROC 6.0).

Alternatively, an attacker might employ a \textbf{Universal Perturbation Attack}, seeking a single perturbation applicable to any image class (real/fake). While universal perturbations are less
effective than ensemble or diverse image-specific attacks, they still pose a significant threat as they can be added to an image on the fly,
without any computation, and can be shared easily. They achieve an average AUROC of 36.8, outperforming \ac{PGD} transfer attacks (48.8 AUROC), a trend observed across all the datasets. In contrast, a \textbf{Query-efficient Attack} is one of the least effective in our setup. We report an average AUROC of 64.9. We attribute this lower performance to the restricted query budget and sampling strategy.

Overall, our findings show that \ac{PGD} attacks have limited transferability. This is because they heavily rely on the specific gradients of the source model. Ensemble-based attacks significantly improve adversarial effectiveness, especially when combined with input diversity. This discourages overfitting to any single model and makes the attack robust to common image transformations, ensuring the perturbation is not reliant on specific pixel locations. Although universal perturbations are only computed once, they offer a modest improvement over standard \ac{PGD} and are far easier to deploy, making them a scalable attack. In contrast, query-efficient attacks remain the least effective in our evaluation and appear less promising under current constraints.

\begin{insight}
Standard \ac{PGD} attacks struggle to transfer across different architectures and models. Ensembling detectors, particularly when augmented with input diversity, produces highly effective and transferable adversarial examples.
\end{insight}

\begin{figure*}[t]
    \centering
    \includegraphics[width=0.9\textwidth]{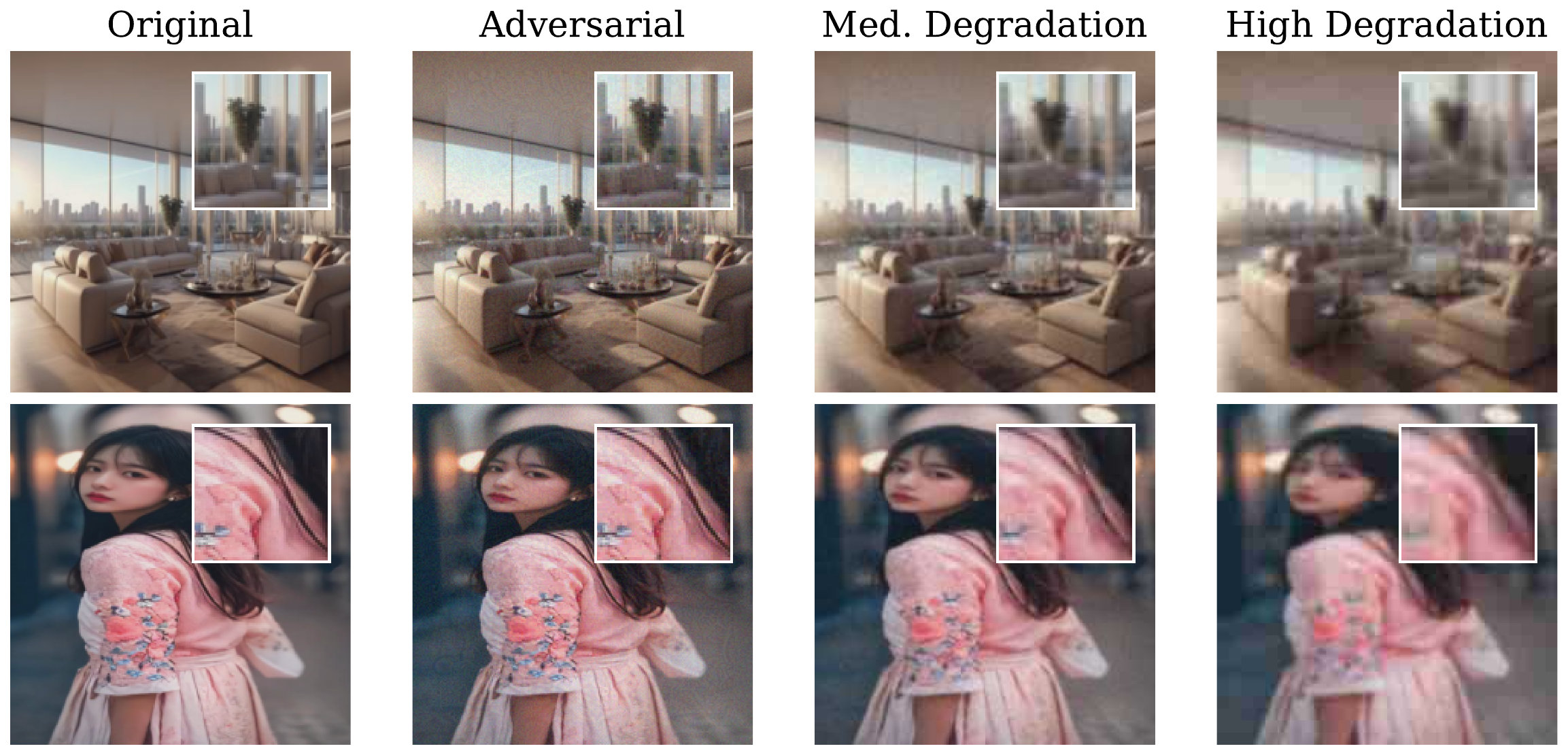}
    \caption{\textbf{Visual effect of various degradation levels on adversarial examples.} Example showing an adversarial example (Diverse Input attack, $\epsilon = 8/255$) and its appearance after medium and high degradation (degraded adversarial). Robust attacks aim to remain effective after such transformations.}
    \label{fig:combined_degradation}
\end{figure*}

\subsection{Attacks in Real-world Scenarios}
While the previous analysis demonstrates attack effectiveness without degradation, real-world scenarios introduce additional complexities. We consider attacks in a setting, where adversarial examples, spread via social media or messaging apps, frequently undergo transformations such as resizing and compression, as shown by Xu \etal~\cite{xuProFakeDetectingDeepfakes2024}. Our analysis assumes that an attacker aims to evade detection across various degradation levels (none, medium, high) encountered in an image's digital life cycle, rather than focusing on less common, specific edits. To illustrate the visual impact of these settings, Figure~\ref{fig:combined_degradation} provides a qualitative comparison between the original image, the adversarial examples, and their degraded counterparts.
To ensure a fair comparison and measure the effectiveness of the attack without bias, we first report detector performance in the clean setting over these degradations in Table~\ref{tab:attack_results_averaged}.
The results show that most detectors are generally robust to these transformations, and their performance is slightly 
affected. CLIP-based detectors (UnivFD and DRCT-CLIP) exhibit greater robustness, potentially due to their reliance on high-level 
features. In contrast, CNN-based detectors (DRCT-ConvB and Corvi), which depend more on low-level features, are consequently more
affected by these (often low-level) transformations, even though such common transformations are part of the training set for all
detectors.

\begin{figure*}[t]
    \centering
    \begin{subfigure}[t]{0.48\textwidth}
        \includegraphics[width=\textwidth]{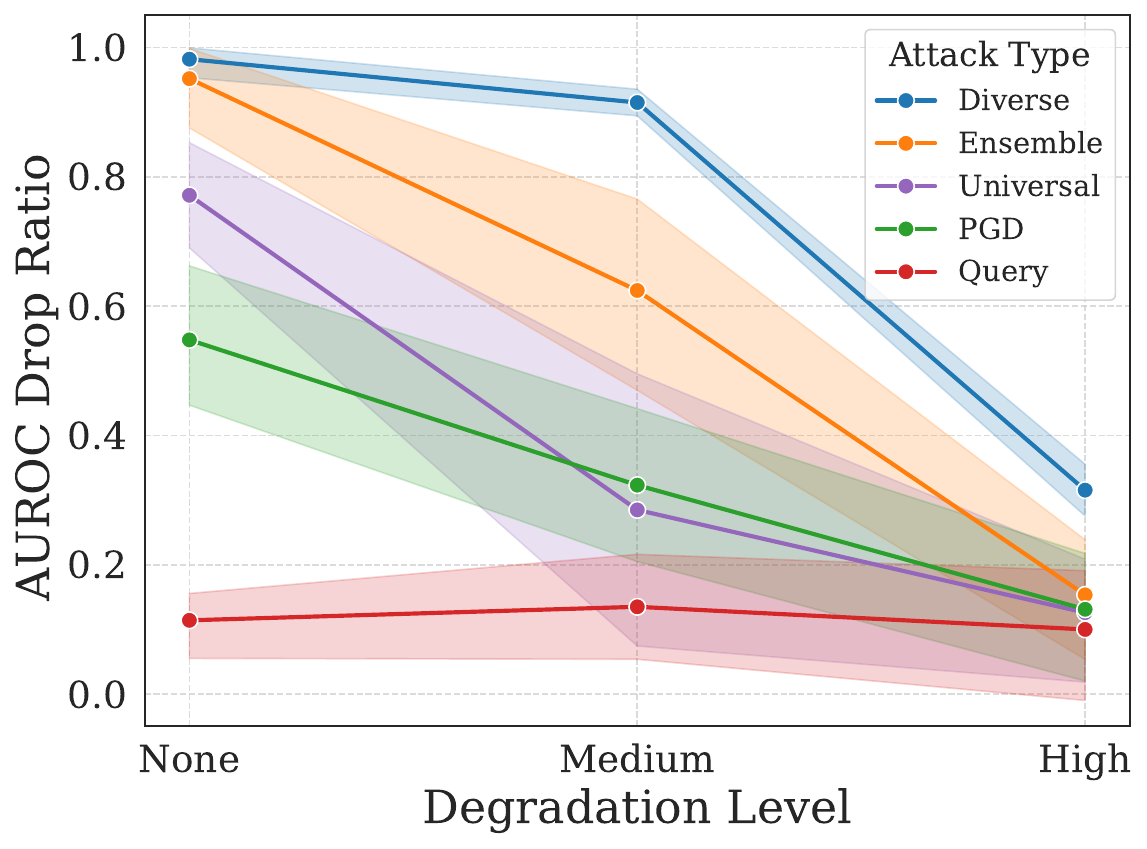}
        \caption{}
        \label{fig:degradation_effect}
    \end{subfigure}
    \hfill
    \begin{subfigure}[t]{0.48\textwidth}
        \includegraphics[width=\textwidth]{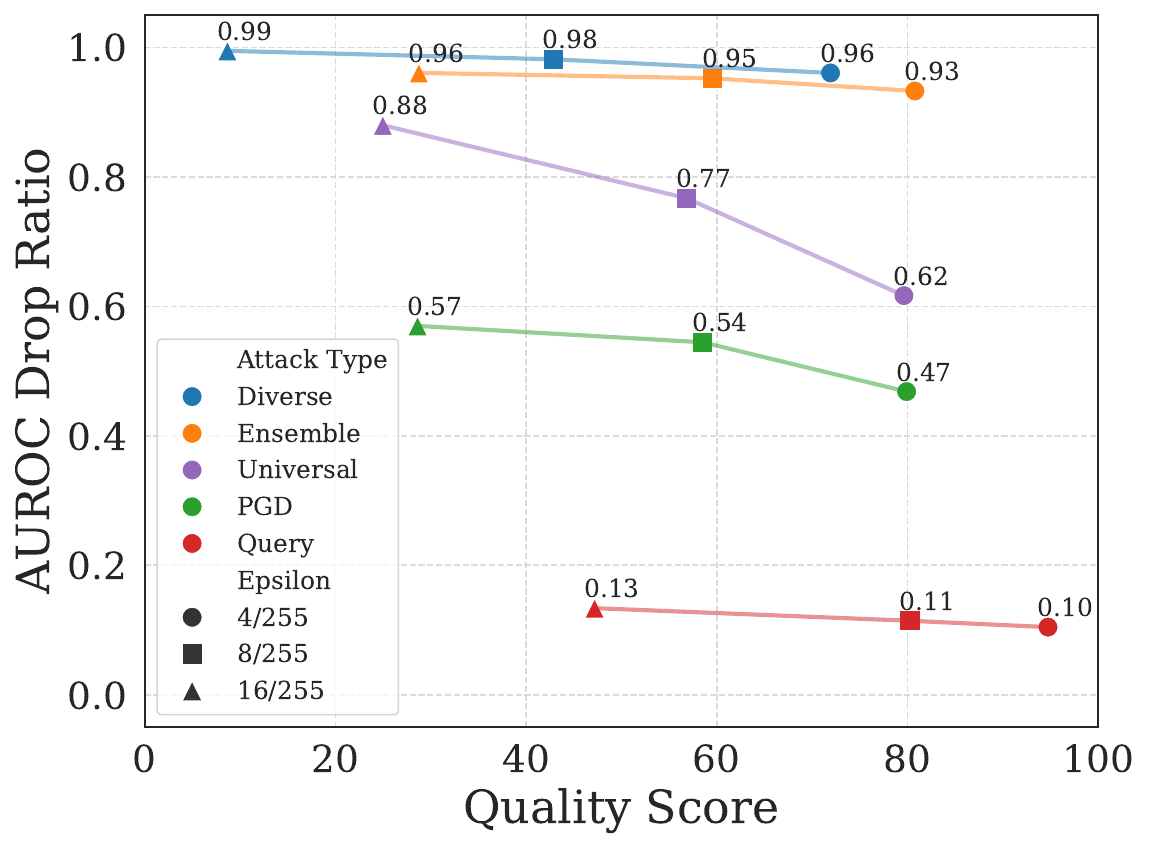}
        \caption{}
        \label{fig:qual_asr}
    \end{subfigure}
    \caption{\textbf{Impact of Degradation on Attack Effectiveness and Quality.} (a) Impact of degradation severity on attack effectiveness, quantified by the AUROC Drop Ratio.
     Results are averaged over all tested detectors and datasets.
     (b) Comparison of three $\epsilon$ for tested attacks.
     Image quality decreases notably with larger $\epsilon$ (lower quality score), while AUROC remains relatively stable, suggesting that minimal perturbations suffice for effective attacks with high image quality.}
    \label{fig:qual_asr_combined}
\end{figure*}

Subsequently, we evaluate the impact of image degradations on the effectiveness of adversarial examples. The overall impact on detector performance is detailed in the right two columns of Table~\ref{tab:attack_results_averaged} and further illustrated in \Cref{fig:degradation_effect}.

In general, most attacks become less effective after transformation, although they can still pose a substantial threat. The \textbf{Diverse Input attack} shows greater robustness to degradation compared to the \textbf{Ensemble attack}, as evidenced by a smaller performance drop from the ``None'' to ``Medium'' degradation levels. We hypothesize that this is due to the incorporation of resizing into the adversarial example generation process, which helps the attack generalize better under transformation. In contrast, the \textbf{Universal attack} is the least robust, suffering from a significant performance decline when comparing no degradation to medium degradation (as showcased by the drop in \Cref{fig:degradation_effect}). This is expected, as universal perturbations are generic and intended to work across an entire dataset; even minor alterations can make them ineffective.

In summary, while transformations reduce attack effectiveness (as indicated by the AUROC drop ratio), even severe degradations that significantly alter the image do not completely neutralize the adversarial noise.  It is also noteworthy that medium degradation is the setting closest to common social-media processing, following the degradation model of Xu~\etal~\cite{xuProFakeDetectingDeepfakes2024}, while high degradation captures more extreme cases such as aggressive compression or repeated post-processing.

\begin{insight}
While degradations on adversarial examples reduce attack effectiveness, they do not neutralize them. Diverse Input attacks are most robust to these transformations, while Universal attacks are most fragile.
\end{insight}

\begin{table}[t]
    \centering
    \caption{\textbf{Quantitative evaluation of various attack methods.} The table compares five attacks using image quality metrics and the AUROC drop ratio as an attack-effectiveness measure. Adversarial examples are generated with $\epsilon=8/255$ in the no-degradation black-box setting. The arrows ($\uparrow/\downarrow$) indicate the preferable direction for each metric. %
}
    \footnotesize
    \setlength{\tabcolsep}{4pt}
    \sisetup{detect-weight=true, detect-inline-weight=math}
    \resizebox{.98\columnwidth}{!}{
    \begin{tabular}{@{\hspace{2pt}}l *{5}{S[table-format=2.3]}@{\hspace{2pt}}}
        \toprule
        \textbf{Metric} (Range) & {\textbf{PGD}} & {\textbf{Ensemble}} & {\textbf{Diverse}} & {\textbf{Universal}} & {\textbf{Query}} \\
        \midrule
        Mean Pert.~$\downarrow$ [0, $\infty$) & {0.077} & {0.076} & {0.108} & {0.020} & {0.058} \\
        SSIM~$\uparrow$ [0, 1] & {0.917} & {0.921} & {0.871} & {0.875} & {0.943} \\
        PSNR~$\uparrow$ [0, $\infty$) & {33.97} & {34.07} & {31.63} & {32.48} & {49.52} \\
        LPIPS~$\downarrow$ [0, $\infty$) & {0.093} & {0.092} & {0.161} & {0.156} & {0.053} \\
        DISTS~$\downarrow$ [0, $\infty$) & {0.127} & {0.122} & {0.165} & {0.140} & {0.079} \\
        \midrule %
        Quality Score~$\uparrow$ [0, 100] & {58.12} & {59.27} & {40.73} & {55.14} & { 81.82} \\
        AUROC Drop Ratio~$\uparrow$ [0, 100] & {54.2} & {95.0} & {98.0} & {76.4} & {11.4} \\
        \bottomrule
    \end{tabular}
    }
\label{tab:attack-quality}
\end{table}

\subsection{Adversarial Example Quality Analysis}
Analyzing the quality of adversarial examples is important to ensure they are not only effective in evading detection but also maintain high visual fidelity.

\paragraph{Setup} To quantitatively evaluate the quality of adversarial examples, we employ five widely-used metrics: Mean Perturbation, Peak Signal-to-Noise Ratio (PSNR), Structural Similarity Index (SSIM), Learned Perceptual Image Patch Similarity (LPIPS)~\cite{lpips}, and Deep Image Structure and Texture Similarity (DISTS)~\cite{dist-quality}, selected to capture different aspects of image quality, including perceptual similarity and structural
integrity. These metrics are computed between adversarial examples and their original counterparts without degradation, as constant
degradation applied to both would not alter their relative quality. For a unified and holistic measure, we compute a normalized average of these
five metrics with similar weights, presented as ``Quality'' (0-100, with 100 being the highest similarity) in Table~\ref{tab:attack-quality}. %

\paragraph{Quality-Attack Success Trade-off}
The results in Table~\ref{tab:attack-quality} reveal a strong correlation between an attack's effectiveness, measured by AUROC drop ratio, and the visual quality of the generated adversarial examples. For instance, the Ensemble attack and Universal attack demonstrate quality on par with the \ac{PGD} attack, yet their effectiveness scores are substantially higher (0.95 for Ensemble and 0.76 for Universal, compared to 0.54 for \ac{PGD}), indicating greater impact while preserving similar visual fidelity. Conversely, the Diverse Input attack, which has the highest score of 0.98, exhibits the lowest quality. To facilitate a fair comparison, we plot this success metric against the Quality Score for three distinct $\epsilon$ values (4/255, 8/255, and 16/255) in \Cref{fig:qual_asr}. This analysis shows a significant decrease in quality with increasing $\epsilon$, while the corresponding gain in attack efficacy is not as pronounced for methods like \ac{PGD}, Ensemble, Diverse Input, and Query-efficient. This demonstrates that an attacker can achieve high impact by carefully selecting the attack method and an appropriate $\epsilon$ while maintaining excellent visual fidelity. Samples of these adversarial example images are shown in Figure~\ref{fig:qual} in the Appendix.

\begin{insight}
The trade-off between image quality and attack success is not linear. Increasing perturbation magnitude ($\epsilon$) significantly degrades quality without proportionally increasing attack success. This shows that the optimization technique matters more than the perturbation budget.
\end{insight}

\subsection{Case Study -- Commercial Detector }

Our analysis identifies the Diverse Input attack as highly effective, achieving a significant AUROC drop ratio while producing good quality images with an appropriate epsilon value.
To assess its real-world applicability, we conducted a case study on HIVE~\cite{hive}, a widely-used closed-source online \ac{AIGI} detector. Using HIVE's free web interface, we tested the Diverse Input attack (using an ensemble of all four detectors) on a subset of 200 images (100 from each of the Synthbuster and Chameleon datasets, representing 20\% of our evaluation set). This sample size was constrained by the interface's rate limits. The results are presented in Table~\ref{tab:case_study}. Notably, in the clean setting, commercial detectors like HIVE significantly outperform their academic counterparts, with HIVE achieving near-perfect AUROC on both datasets. However, when attacked with $\epsilon = 8/255$, HIVE's AUROC dropped from 100.0 to 73.8 on Synthbuster and from 97.8 to 79.4 on Chameleon dataset. 

\begin{table}[!t]
    \caption{\textbf{Case Study on HIVE.} Performance under attacks showing AUROC, Accuracy, and TPR@5\%FPR (\%) under different perturbation bounds.}
    \centering
    \setlength{\tabcolsep}{6pt}
    \begin{tabular}{@{\hspace{2pt}} l l *{3}{S[table-format=3.1]} @{\hspace{2pt}}}
        \toprule
        \multirow{2}{*}{\textbf{Metric}} & \multirow{2}{*}{\textbf{Dataset}} & & \multicolumn{2}{c}{\textbf{Attacked}} \\
        \cmidrule(lr){4-5}
        & & {\textbf{Clean}} & {$\mathbf{\epsilon=4/255}$} & {$\mathbf{\epsilon=8/255}$} \\
        \midrule
        \multirow{3}{*}{\textbf{AUROC}}
            & Synthbuster & 100.0 & 90.3 & 73.8 \\[0.2ex]
            & Chameleon & 97.8 & 83.3 & 79.4 \\[0.2ex]
            \cmidrule{2-5}
            & Average & 98.9 & 86.8 & 76.6 \\[0.8ex]
        \midrule
        \multirow{3}{*}{\textbf{Accuracy}}
            & Synthbuster & 100.0 & 82.0 & 65.0 \\[0.2ex]
            & Chameleon & 91.0 & 77.0 & 75.7 \\[0.2ex]
            \cmidrule{2-5}
            & Average & 95.5 & 79.5 & 70.4 \\[0.8ex]
        \midrule
        \multirow{3}{*}{\textbf{TPR@5\%FPR}}
            & Synthbuster & 100.0 & 72.0 & 30.0 \\[0.2ex]
            & Chameleon & 98.0 & 52.0 & 30.0 \\[0.2ex]
            \cmidrule{2-5}
            & Average & 99.0 & 62.0 & 30.0 \\
        \bottomrule
    \end{tabular}
    \label{tab:case_study}
\end{table}

To conduct a more in-depth analysis, we used the 0.9 threshold suggested by HIVE and measured the attack performance on real and fake classes separately. A key observation is the divergent impact of adversarial perturbations across datasets. On Chameleon, the attack primarily induces false positives. Comparing perturbation magnitudes reveals that at $\epsilon=4/255$, fake detection accuracy drops to 72.0\%, but at $\epsilon=8/255$, it recovers to 82.0\%, while real accuracy continues to degrade (from 82.0\% to 69.4\%). This suggests that for the high-resolution images in Chameleon, high-magnitude adversarial noise acts as an additive artifact that mimics synthetic traits.
Upon disclosure, the HIVE team confirmed the validity of our results.

\begin{table}[t]
    \centering
    \caption{\textbf{Defense Performance.} AUROC, Accuracy, and TPR@5\%FPR for robust models (UnivFD-R2, UnivFD-R4) compared to baseline UnivFD, averaged across Synthbuster, Chameleon, and GPT-4o.}
    \footnotesize
    \setlength{\tabcolsep}{2pt}
    \resizebox{0.9\columnwidth}{!}{
    \begin{tabular}{@{\hspace{2pt}}ll|ccc|ccc|ccc@{\hspace{2pt}}}
        \toprule
        \multirow{2}{*}{\textbf{Attack}} & \multirow{2}{*}{\textbf{Target}} & \multicolumn{3}{c|}{\textbf{AUROC}} & \multicolumn{3}{c|}{\textbf{Accuracy}} & \multicolumn{3}{c}{\textbf{TPR@5\%FPR}} \\
        \cmidrule(lr){3-5} \cmidrule(lr){6-8} \cmidrule(lr){9-11}
        && \textbf{None} & \textbf{Med} & \textbf{High} & \textbf{None} & \textbf{Med} & \textbf{High} & \textbf{None} & \textbf{Med} & \textbf{High} \\
        \midrule
        \multirow{3}{*}{\textbf{Clean}} & UnivFD & 91.4 & 89.5 & 89.7 & 80.9 & 77.7 & 82.7 & 65.9 & 60.9 & 66.4 \\
         & UnivFD-R2 & 88.3 & 85.8 & 84.1 & 82.7 & 77.8 & 77.1 & 62.4 & 53.9 & 48.1 \\
         & UnivFD-R4 & 84.4 & 82.6 & 81.1 & 77.7 & 76.3 & 74.9 & 54.4 & 51.1 & 45.3 \\
        \midrule
        \midrule
        \multirow{3}{*}{\textbf{PGD}} & UnivFD & 0.1 & 34.4 & 58.6 & 2.4 & 41.5 & 53.9 & 0.0 & 3.7 & 11.6 \\
         & UnivFD-R2 & 72.3 & 78.3 & 74.1 & 66.0 & 72.0 & 66.3 & 31.1 & 38.5 & 28.4 \\
         & UnivFD-R4 & 76.5 & 78.1 & 74.9 & 69.9 & 70.2 & 67.3 & 33.3 & 39.2 & 25.7 \\
        \midrule
        \multirow{3}{*}{\textbf{Ensemble}} & UnivFD & 0.0 & 51.5 & 63.7 & 0.6 & 52.5 & 54.7 & 0.0 & 8.1 & 14.3 \\
         & UnivFD-R2 & 75.8 & 81.3 & 75.2 & 68.6 & 74.8 & 67.7 & 36.7 & 45.2 & 32.3 \\
         & UnivFD-R4 & 78.1 & 79.3 & 75.3 & 70.5 & 71.4 & 67.3 & 37.9 & 39.2 & 26.0 \\
        \midrule
        \multirow{3}{*}{\textbf{Diverse}} & UnivFD & 0.1 & 6.0 & 47.9 & 3.1 & 19.4 & 50.3 & 0.0 & 0.0 & 6.1 \\
         & UnivFD-R2 & 60.5 & 75.7 & 72.8 & 56.9 & 68.3 & 65.2 & 16.9 & 38.7 & 30.3 \\
         & UnivFD-R4 & 72.2 & 77.2 & 74.5 & 65.9 & 69.2 & 66.9 & 29.7 & 35.9 & 25.5 \\
        \midrule
        \multirow{3}{*}{\textbf{Universal}} & UnivFD & 28.6 & 79.4 & 65.3 & 35.6 & 69.1 & 57.6 & 0.8 & 34.8 & 16.3 \\
         & UnivFD-R2 & 77.3 & 82.3 & 75.4 & 68.3 & 72.9 & 67.1 & 36.1 & 45.6 & 31.1 \\
         & UnivFD-R4 & 78.1 & 79.8 & 75.4 & 70.3 & 71.0 & 67.5 & 36.4 & 39.3 & 27.2 \\
        \midrule
        \multirow{3}{*}{\textbf{Query}} & UnivFD & 66.2 & 76.6 & 68.7 & 59.1 & 67.6 & 57.2 & 21.3 & 36.8 & 18.1 \\
         & UnivFD-R2 & 76.5 & 81.2 & 75.6 & 69.7 & 74.3 & 67.9 & 38.4 & 45.9 & 30.9 \\
         & UnivFD-R4 & 78.2 & 79.4 & 75.6 & 70.8 & 71.9 & 67.6 & 37.5 & 40.3 & 28.0 \\
        \bottomrule
    \end{tabular}
    }
    \label{tab:defense_results_avg}
\end{table}

\subsection{Towards Defending AIGI Detectors}
\label{sec:defense}

Given the demonstrated vulnerabilities, we explore the potential of leveraging robustly pre-trained feature extractors as a defense mechanism. It is important to emphasize that our objective here is \emph{not} to present this as the \textit{definitive} or \textit{optimal} defense. Rather, our aim is to provide a preliminary showcase: to investigate whether established defense paradigms can offer a degree of robustness in this rapidly evolving landscape and to stimulate further research into tailored \ac{AIGI} adversarial robustness.

\paragraph{Defense Method}
We adapt an existing defense based on robust feature extractors to \ac{AIGI} detection. Our approach builds on RobustCLIP~\cite{schlarmann2024robustclip}, a version of CLIP fine-tuned with unsupervised adversarial training to produce perturbation-invariant feature representations. This choice is motivated by two observations: (1) CLIP-based detectors (UnivFD, DRCT-CLIP) achieve strong detection and generalization performance, and (2) robust pre-trained backbones can transfer adversarial robustness to downstream tasks~\cite{transfer-rboust-24}.

\paragraph{Architecture and Training}
We replace the standard CLIP-ViT-L/14 backbone in UnivFD with the RobustCLIP backbone. The architecture remains otherwise identical: a frozen backbone followed by a single linear classification layer. We train only the linear head using the same training data (GenImage SD-1.4 subset) and hyperparameters as the original UnivFD model. The key difference from the baseline is the use of the adversarially-trained backbone, which provides robustness without requiring adversarial training of the detector itself.

We evaluate two variants: \textbf{UnivFD-R2} and \textbf{UnivFD-R4}, which use RobustCLIP backbones trained with $\epsilon = 2/255$ and $\epsilon = 4/255$, respectively. Higher $\epsilon$ during backbone training generally yields stronger robustness but may reduce clean performance due to the robustness-accuracy trade-off.

\paragraph{Evaluating the Defended Models}
The defended models are evaluated against the same attacks and degradations used in previous experiments with  $ \epsilon = 8/255$. 
Table~\ref{tab:defense_results_avg} reports AUROC averaged across all datasets on defended and reference models. The defended models demonstrate improved robustness against all attacks: the AUROC score of the undefended model drops to nearly 0 for our strongest attack (Diverse Input) in the clean setting, while the defended models achieve AUROC scores of $68.9$ (UnivFD-R2) and $75.7$ (UnivFD-R4). 
This enhanced robustness comes with a slight decrease in clean AUROC, from 91.4 for the undefended UnivFD to 88.3 (R2) and 84.4 (R4).
Similar improvements are observed for attacks combined with various levels of degradation; however, we do not nearly reach the same performance as in the clean case. The defended models achieve AUROC of 78.8 (R2) and 78.3 (R4), compared to 6.7 for the undefended UnivFD. Although UnivFD-R4 demonstrates a higher degree of robustness, detector deployers must consider the detector's susceptibility to attacks and the inherent trade-off between robustness and clean performance. 

\begin{insight}
Replacing standard backbones with robustly pre-trained feature extractors can improve robustness against strong attacks, but comes with a 3.1--7.0 point drop in clean AUROC. %
\end{insight}

\section{Discussion}

Our results show that state-of-the-art \ac{AIGI} detectors are highly sensitive to realistic degradations and adversarial perturbations, even when images remain visually convincing to humans. In this section, we discuss what these findings imply for how \ac{AIGI} detection should be defined, evaluated, and deployed in practice, as well as the limitations of our research.

\paragraph{Detection vs.\ Generation Arms Race}
Preventing misuse of \ac{AIGI} is an arms race between AI generation techniques and detection mechanisms. As soon as a new detection method is developed, generative techniques are devised to circumvent these detectors, propagating a continuous cycle of advancements on both sides. This dynamic nature makes developing persistent solutions for AI-generated content detection challenging.
In the future, methods must take into account that detection cannot be perfect, but we must also study the limitations of detection methods. Our work offers one step toward defining and evaluating realistic threat models and measuring their robustness against attacks.

\paragraph{Context-dependent Detection} AI-generated content is becoming an integral part of everyday life, with its usage expected to rise significantly. As a result, auditing every AI-generated image may become impractical. Instead, the focus should shift toward detecting harmful AI-generated content to identify and mitigate associated risks. For instance, in social media, AI-generated images are more likely to be used in disinformation and scams.
Therefore, detection methods should concentrate on the most critical scenarios and consider characteristics unique to these contexts.

\paragraph{Robustness-Accuracy Trade-off} Enhancing forensic classifier robustness often reduces accuracy on clean data. Deploying these systems requires weighing this trade-off based on the likelihood of attacks. Balancing robustness and accuracy demands careful, context-aware evaluation. In some cases, utility may outweigh full AI-content detection (e.g., less interactive platforms), while others require catching every instance (e.g., news sites). Thus, classification needs must be accurately assessed to align performance with specific use cases.

\paragraph{Limitations}
Our study has several limitations.
First, we focus on a subset of generators, datasets, and post-processing pipelines, so our evaluation should be viewed as covering representative but not exhaustive scenarios.
Second, our commercial case study on HIVE is small-scale due to the web interface limitation. Although it illustrates a concrete instance of fragility in real-world systems, it does not provide a full characterization of real-world risk across detection platforms.
Finally, our defense analysis is scoped to a limited set of detectors, so future work should compare a broader range of robustness techniques under the same threat models.

\section{Conclusion}
Our evaluation reveals that state-of-the-art AIGI detectors remain highly vulnerable to adversarial attacks. Our findings indicate that adversarial examples transfer across different detectors, posing a significant threat to forensic detection mechanisms. Moreover, real-world degradations, such as those induced by social media post-processing, do not render well-chosen attacks ineffective.

Our work is one step toward defining and evaluating realistic threat models, and we argue that future research must accurately assess classification needs to align detector performance with specific, critical use cases. Our results on using robust features show that while the robustness of detectors can be improved, they do not achieve the same performance as in clean settings. We argue that to prevent the misuse of GenAI, the field requires clear definitions, better risk assessments, and alternative solutions that consider the context of misuse.

\section{Ethical Considerations}
\label{sec:ethics}

This research is dual-use, as our attack methods could be misused to evade detectors. We practiced responsible disclosure by informing HIVE before publishing. We believe technical solutions alone are insufficient; a broader strategy integrating technology with education and policy is essential.

\section{Use of Generative AI}
GitHub Copilot assisted in coding and debugging experimental scripts. LLMs helped with polishing the manuscript for clarity. All outputs were independently verified by the authors. The research used existing third-party datasets of AI-generated images.

\section*{Acknowledgments}
This work was supported by the German Federal Ministry of Education and Research under the grants AIgenCY (16KIS2012) and SisWiss (16KIS2330). Moreover, it was funded by the Deutsche Forschungsgemeinschaft (DFG, German Research Foundation) under Germany's Excellence Strategy - EXC 2092 CASA - 390781972, the LCIS center VW-Vorab-2025, ZN4704 11-76251-2055, as well as the Daimler and Benz Foundation under the grant Ladenburger Kolleg, Project KonCheck.

\bibliographystyle{splncs04}
\bibliography{bib}

@inproceedings{abdullah-24-analysis,
	title        = {An Analysis of Recent Advances in Deepfake Image Detection in an Evolving Threat Landscape},
	author       = {Abdullah, Sifat Muhammad and Cheruvu, Aravind and Kanchi, Shravya and Chung, Taejoong and Gao, Peng and Jadliwala, Murtuza and Viswanath, Bimal},
	year         = 2024,
	booktitle    = {IEEE Symposium on Security and Privacy (S\&P)}
}

@misc{c2pa,
	title        = {{C2PA} Technical Specification},
	author       = {{Coalition for Content Provenance and Authenticity (C2PA)}},
	year         = 2024,
	note         = {\url{https://c2pa.org/specifications/specifications/2.0/specs/C2PA_Specification.html}}
}

@inproceedings{carlini_2020,
	title        = {Evading Deepfake-Image Detectors with White- and Black-Box Attacks},
	author       = {Carlini, Nicholas and Farid, Hany},
	year         = 2020,
	booktitle    = {IEEE/CVF Conference on Computer Vision and Pattern Recognition Workshops (CVPRW)},
}

@inproceedings{clip-21,
	title        = {Learning Transferable Visual Models from Natural Language Supervision},
	author       = {Radford, Alec and Kim, Jong Wook and Hallacy, Chris and Ramesh, Aditya and Goh, Gabriel and Agarwal, Sandhini and Sastry, Girish and Askell, Amanda and Mishkin, Pamela and Clark, Jack and Krueger, Gretchen and Sutskever, Ilya},
	year         = 2021,
	booktitle    = {International Conference on Machine Learning (ICML)},
}

@inproceedings{cnnspot-20,
	title        = {{CNN}-Generated Images Are Surprisingly Easy to Spot... for Now},
	author       = {Wang, Sheng-Yu and Wang, Oliver and Zhang, Richard and Owens, Andrew and Efros, Alexei A.},
	year         = 2020,
	booktitle    = {IEEE/CVF Conference on Computer Vision and Pattern Recognition (CVPR)}
}

@inproceedings{corviDetectionSyntheticImages2023,
	title        = {On the Detection of Synthetic Images Generated by Diffusion Models},
	author       = {Corvi, Riccardo and Cozzolino, Davide and Zingarini, Giada and Poggi, Giovanni and Nagano, Koki and Verdoliva, Luisa},
	year         = 2023,
	booktitle    = {IEEE International Conference on Acoustics, Speech and Signal Processing (ICASSP)}
}

@inproceedings{Cozzolino_2024_CVPR,
	title        = {Raising the Bar of {AI}-Generated Image Detection with {CLIP}},
	author       = {Cozzolino, Davide and Poggi, Giovanni and Corvi, Riccardo and Nie{\ss}ner, Matthias and Verdoliva, Luisa},
	year         = 2024,
	booktitle    = {IEEE/CVF Conference on Computer Vision and Pattern Recognition (CVPR) Workshops}
}

@article{dist-quality,
	title        = {Image Quality Assessment: Unifying Structure and Texture Similarity},
	author       = {Ding, Keyan and Ma, Kede and Wang, Shiqi and Simoncelli, Eero P.},
	year         = 2022,
	journal      = {IEEE Transactions on Pattern Analysis and Machine Intelligence},
}

@inproceedings{dosovitskiyImageWorth16x162020,
	title        = {An Image Is Worth 16x16 Words: Transformers for Image Recognition at Scale},
	author       = {Dosovitskiy, Alexey and Beyer, Lucas and Kolesnikov, Alexander and Weissenborn, Dirk and Zhai, Xiaohua and Unterthiner, Thomas and Dehghani, Mostafa and Minderer, Matthias and Heigold, Georg and Gelly, Sylvain and Uszkoreit, Jakob and Houlsby, Neil},
	year         = 2020,
	booktitle    = {International Conference on Learning Representations (ICLR)}
}

@inproceedings{drct-24,
	title        = {{DRCT}: Diffusion Reconstruction Contrastive Training towards Universal Detection of Diffusion Generated Images},
	author       = {Chen, Baoying and Zeng, Jishen and Yang, Jianquan and Yang, Rui},
	year         = 2024,
	booktitle    = {International Conference on Machine Learning (ICML)}
}

@inproceedings{fei-22-supervised,
	title        = {Supervised {GAN} Watermarking for Intellectual Property Protection},
	author       = {Fei, Jianwei and Xia, Zhihua and Tondi, Benedetta and Barni, Mauro},
	year         = 2022,
	booktitle    = {IEEE International Workshop on Information Forensics and Security (WIFS)}
}

@inproceedings{fernandez-23-stable,
	title        = {The Stable Signature: Rooting Watermarks in Latent Diffusion Models},
	author       = {Fernandez, Pierre and Couairon, Guillaume and J{\'e}gou, Herv{\'e} and Douze, Matthijs and Furon, Teddy},
	year         = 2023,
	booktitle    = {IEEE/CVF International Conference on Computer Vision (ICCV)}
}

@misc{flux,
	title        = {{FLUX}},
	author       = {{Black Forest Labs}},
	year         = 2024,
	note         = {\url{https://github.com/black-forest-labs/flux}}
}

@inproceedings{frank_2023,
	title        = {A Representative Study on Human Detection of Artificially Generated Media Across Countries},
	author       = {Frank, Joel and Herbert, Franziska and Ricker, Jonas and Sch{\"o}nherr, Lea and Eisenhofer, Thorsten and Fischer, Asja and D{\"u}rmuth, Markus and Holz, Thorsten},
	year         = 2024,
	booktitle    = {IEEE Symposium on Security and Privacy (S\&P)}
}

@article{genimage-23,
	title        = {{GenImage}: A Million-Scale Benchmark for Detecting {AI}-Generated Image},
	author       = {Zhu, Mingjian and Chen, Hanting and Yan, Qiangyu and Huang, Xudong and Lin, Guoqing and Li, Wei and Tu, Zhijian and Hu, Hailin and Hu, Jie and Wang, Yunhe},
	year         = 2023,
	journal      = {Advances in Neural Information Processing Systems (NeurIPS)}
}

@inproceedings{grag-21,
	title        = {Are {GAN} Generated Images Easy to Detect? A Critical Analysis of the State-of-the-Art},
	author       = {Gragnaniello, Diego and Cozzolino, Davide and Marra, Francesco and Poggi, Giovanni and Verdoliva, Luisa},
	year         = 2021,
	booktitle    = {IEEE International Conference on Multimedia and Expo (ICME)}
}

@misc{hive,
	title        = {{HIVE} Moderation},
	author       = {{HIVE}},
	year         = 2025,
	note         = {\url{https://hivemoderation.com/}}
}

@inproceedings{hou_2023,
	title        = {Evading Deepfake Detectors via Adversarial Statistical Consistency},
	author       = {Hou, Yang and Guo, Qing and Huang, Yihao and Xie, Xiaofei and Ma, Lei and Zhao, Jianjun},
	year         = 2023,
	booktitle    = {IEEE/CVF Conference on Computer Vision and Pattern Recognition (CVPR)},
}

@inproceedings{hu2025a,
	title        = {A Transfer Attack to Image Watermarks},
	author       = {Hu, Yuepeng and Jiang, Zhengyuan and Guo, Moyang and Gong, Neil Zhenqiang},
	year         = 2025,
	booktitle    = {International Conference on Learning Representations (ICLR)}
}

@inproceedings{ilyas2018queryefficientblackboxadversarialexamples,
	title        = {Black-Box Adversarial Attacks with Limited Queries and Information},
	author       = {Ilyas, Andrew and Engstrom, Logan and Athalye, Anish and Lin, Jessy},
	year         = 2018,
	booktitle    = {International Conference on Machine Learning (ICML)},
}

@inproceedings{jiang-23-evading,
	title        = {Evading Watermark Based Detection of {AI}-Generated Content},
	author       = {Jiang, Zhengyuan and Zhang, Jinghuai and Gong, Neil Zhenqiang},
	year         = 2023,
	booktitle    = {ACM SIGSAC Conference on Computer and Communications Security (CCS)}
}

@inproceedings{li_2021,
	title        = {Exploring Adversarial Fake Images on Face Manifold},
	author       = {Li, Dongze and Wang, Wei and Fan, Hongxing and Dong, Jing},
	year         = 2021,
	booktitle    = {IEEE/CVF Conference on Computer Vision and Pattern Recognition (CVPR)},
}

@article{liu_2023,
	title        = {Making Deepfakes More Spurious: Evading Deep Face Forgery Detection via Trace Removal Attack},
	author       = {Liu, Chi and Chen, Huajie and Zhu, Tianqing and Zhang, Jun and Zhou, Wanlei},
	year         = 2023,
	journal      = {IEEE Transactions on Dependable and Secure Computing},
}

@inproceedings{lpips,
	title        = {The Unreasonable Effectiveness of Deep Features as a Perceptual Metric},
	author       = {Zhang, Richard and Isola, Phillip and Efros, Alexei A. and Shechtman, Eli and Wang, Oliver},
	year         = 2018,
	booktitle    = {IEEE/CVF Conference on Computer Vision and Pattern Recognition (CVPR)},
}

@misc{FBI2025IC3,
	title        = {Cryptocurrency and {AI} Scams Bilk Americans of Billions},
	author       = {{Federal Bureau of Investigation (FBI)}},
	year         = 2026,
	note         = {\url{https://www.fbi.gov/news/press-releases/cryptocurrency-and-ai-scams-bilk-americans-of-billions}}
}

@misc{midjourney_24,
	title        = {Midjourney},
	author       = {{Midjourney}},
	year         = 2024,
	note         = {\url{https://www.midjourney.com/home}}
}

@inproceedings{moosavidezfooli2017universaladversarialperturbations,
	title        = {Universal Adversarial Perturbations},
	author       = {Moosavi-Dezfooli, Seyed-Mohsen and Fawzi, Alhussein and Fawzi, Omar and Frossard, Pascal},
	year         = 2017,
	booktitle    = {IEEE/CVF Conference on Computer Vision and Pattern Recognition (CVPR)}
}

@inproceedings{pgd-19,
	title        = {Towards Deep Learning Models Resistant to Adversarial Attacks},
	author       = {Madry, Aleksander and Makelov, Aleksandar and Schmidt, Ludwig and Tsipras, Dimitris and Vladu, Adrian},
	year         = 2018,
	booktitle    = {International Conference on Learning Representations (ICLR)}
}

@inproceedings{raise-15,
	title        = {{RAISE}: A Raw Images Dataset for Digital Image Forensics},
	author       = {Dang-Nguyen, Duc-Tien and Pasquini, Cecilia and Conotter, Valentina and Boato, Giulia},
	year         = 2015,
	booktitle    = {ACM Multimedia Systems Conference},
}

@inproceedings{resnet-15,
	title        = {Deep Residual Learning for Image Recognition},
	author       = {He, Kaiming and Zhang, Xiangyu and Ren, Shaoqing and Sun, Jian},
	year         = 2016,
	booktitle    = {IEEE/CVF Conference on Computer Vision and Pattern Recognition (CVPR)},
}

@inproceedings{ricker-twitter-2024,
	title        = {{AI}-Generated Faces in the Real World: A Large-Scale Case Study of {Twitter} Profile Images},
	author       = {Ricker, Jonas and Assenmacher, Dennis and Holz, Thorsten and Fischer, Asja and Quiring, Erwin},
	year         = 2024,
	booktitle    = {International Symposium on Research in Attacks, Intrusions and Defenses (RAID)}
}

@inproceedings{schlarmann2024robustclip,
	title        = {Robust {CLIP}: Unsupervised Adversarial Fine-Tuning of Vision Embeddings for Robust Large Vision-Language Models},
	author       = {Schlarmann, Christian and Singh, Naman Deep and Croce, Francesco and Hein, Matthias},
	year         = 2024,
	booktitle    = {International Conference on Machine Learning (ICML)}
}

@misc{stable-diffusion,
	title        = {{Stable Diffusion}},
	author       = {{Stability AI}},
	year         = 2022,
	note         = {\url{https://stability.ai/stable-image}}
}

@article{synthbuster-2023,
	title        = {Synthbuster: Towards Detection of Diffusion Model Generated Images},
	author       = {Bammey, Quentin},
	year         = 2023,
	journal      = {IEEE Open Journal of Signal Processing},
}

@article{tariang2024syntheticimageverificationera,
	title        = {Synthetic Image Verification in the Era of Generative Artificial Intelligence: What Works and What Isn't There Yet},
	author       = {Tariang, Diangarti and Corvi, Riccardo and Cozzolino, Davide and Poggi, Giovanni and Nagano, Koki and Verdoliva, Luisa},
	year         = 2024,
	journal      = {IEEE Security and Privacy},
}

@inproceedings{tramer2020ensemble,
	title        = {Ensemble Adversarial Training: Attacks and Defenses},
	author       = {Tram{\`e}r, Florian and Kurakin, Alexey and Papernot, Nicolas and Goodfellow, Ian and Boneh, Dan and McDaniel, Patrick},
	year         = 2018,
	booktitle    = {International Conference on Learning Representations (ICLR)}
}

@inproceedings{transfer-rboust-24,
	title        = {Initialization Matters for Adversarial Transfer Learning},
	author       = {Hua, Andong and Gu, Jindong and Xue, Zhiyu and Carlini, Nicholas and Wong, Eric and Qin, Yao},
	year         = 2024,
	booktitle    = {IEEE/CVF Conference on Computer Vision and Pattern Recognition (CVPR)},
}

@inproceedings{univfd-23,
	title        = {Towards Universal Fake Image Detectors that Generalize across Generative Models},
	author       = {Ojha, Utkarsh and Li, Yuheng and Lee, Yong Jae},
	year         = 2023,
	booktitle    = {IEEE/CVF Conference on Computer Vision and Pattern Recognition (CVPR)}
}

@inproceedings{xie2019improvingtransferabilityadversarialexamples,
	title        = {Improving Transferability of Adversarial Examples with Input Diversity},
	author       = {Xie, Cihang and Zhang, Zhishuai and Zhou, Yuyin and Bai, Song and Wang, Jianyu and Ren, Zhou and Yuille, Alan L.},
	year         = 2019,
	booktitle    = {IEEE/CVF Conference on Computer Vision and Pattern Recognition (CVPR)},
}

@inproceedings{xuProFakeDetectingDeepfakes2024,
	title        = {{ProFake}: Detecting Deepfakes in the Wild against Quality Degradation with Progressive Quality-Adaptive Learning},
	author       = {Xu, Huiyu and Wang, Yaopeng and Wang, Zhibo and Ba, Zhongjie and Liu, Wenxin and Jin, Lu and Weng, Haiqin and Wei, Tao and Ren, Kui},
	year         = 2024,
	booktitle    = {ACM SIGSAC Conference on Computer and Communications Security (CCS)},
}

@article{yan2025gptimgevalcomprehensivebenchmarkdiagnosing,
	title        = {{GPT-ImgEval}: A Comprehensive Benchmark for Diagnosing {GPT-4o} in Image Generation},
	author       = {Yan, Zhiyuan and Ye, Junyan and Li, Weijia and Huang, Zilong and Yuan, Shenghai and He, Xiangyang and Lin, Kaiqing and He, Jun and He, Conghui and Yuan, Li},
	year         = 2025,
	journal      = {arXiv preprint arXiv:2504.02782}
}

@inproceedings{yan2025sanitycheckaigeneratedimage,
	title        = {A Sanity Check for {AI}-Generated Image Detection},
	author       = {Yan, Shilin and Li, Ouxiang and Cai, Jiayin and Hao, Yanbin and Jiang, Xiaolong and Hu, Yao and Xie, Weidi},
	year         = 2025,
	booktitle    = {International Conference on Learning Representations (ICLR)}
}

@article{direstaHowSpammersScammers2024,
	title        = {How Spammers and Scammers Leverage {AI}-Generated Images on {Facebook} for Audience Growth},
	author       = {DiResta, Ren{\'e}e and Goldstein, Josh A.},
	year         = 2024,
	journal      = {Harvard Kennedy School Misinformation Review}
}

@article{drolsbach2025characterizingaigeneratedmisinformationsocial,
	title        = {Characterizing {AI}-Generated Misinformation on Social Media},
	author       = {Drolsbach, Chiara and Pr{\"o}llochs, Nicolas},
	year         = 2025,
	journal      = {arXiv preprint arXiv:2505.10266}
}

@misc{ryan-mosleyHowGenerativeAI,
	title        = {How Generative {AI} Is Boosting the Spread of Disinformation and Propaganda},
	author       = {Ryan-Mosley, Tate},
	year         = 2023,
	note         = {\url{https://www.technologyreview.com/2023/10/04/1080801/generative-ai-boosting-disinformation-and-propaganda-freedom-house/}}
}

@article{eddoubi2025raiddatasettestingadversarial,
	title        = {{RAID}: A Dataset for Testing the Adversarial Robustness of {AI}-Generated Image Detectors},
	author       = {Eddoubi, Hicham and Ricker, Jonas and Cocchi, Federico and Baraldi, Lorenzo and Sotgiu, Angelo and Pintor, Maura and Cornia, Marcella and Baraldi, Lorenzo and Fischer, Asja and Cucchiara, Rita and Biggio, Battista},
	year         = 2025,
	journal      = {arXiv preprint arXiv:2506.03988}
}

@inproceedings{park2025chimera,
	title        = {Chimera: Creating Digitally Signed Fake Photos by Fooling Image Recapture and Deepfake Detectors},
	author       = {Park, Seongbin and Vilesov, Alexander and Zhang, Jinghuai and Khalili, Hossein and Tian, Yuan and Kadambi, Achuta and Sehatbakhsh, Nader},
	year         = 2025,
	booktitle    = {USENIX Security Symposium}
}

@misc{verge-nano-banana,
	title        = {Google's {Nano Banana Pro} Generates Excellent Conspiracy Fuel},
	author       = {Hart, Robert and Ricker, Thomas},
	year         = 2025,
	note         = {\url{https://www.theverge.com/report/826003/googles-nano-banana-pro-generates-excellent-conspiracy-fuel}}
}

@inproceedings{sm_2018,
	title        = {Detection of {GAN}-Generated Fake Images over Social Networks},
	author       = {Marra, Francesco and Gragnaniello, Diego and Cozzolino, Davide and Verdoliva, Luisa},
	year         = 2018,
	booktitle    = {IEEE Conference on Multimedia Information Processing and Retrieval (MIPR)},
}

\clearpage

\section*{Appendix}

\begin{table*}[h]
    \centering 
    \footnotesize
    \caption{\textbf{Detectors' Robustness.} AUROC, Accuracy, and TPR@5\%FPR averaged for each datasets for clean (no attack) and different attacks with varying degradation levels (None, Medium: 75\% downsampling and JPEG compression at quality 50, and High: Gaussian blur ($5\times5$, $\mu=1$, $\sigma=1$), 50\% downsampling, JPEG compression at quality 20, and Gaussian noise ($\mu=1$, $\sigma=0.1$)). ``LTO'' means Leave-Target-Out: attacks are optimized on all detectors except the evaluated target.}
    \setlength{\tabcolsep}{4pt}
    \resizebox{.98\textwidth}{!}{
        \begin{tabular}{@{\hspace{2pt}}lll|*{4}{S[table-format=2.1,add-integer-zero]}|*{4}{S[table-format=2.1,add-integer-zero]}|*{4}{S[table-format=2.1,add-integer-zero]}@{\hspace{2pt}}}
            \toprule
            \multirow{4}{*}{\textbf{Attack}} & \multirow{4}{*}{\textbf{Target}} & \multirow{4}{*}{\textbf{Source}} & \multicolumn{4}{c|}{\textbf{Synthbuster}} & \multicolumn{4}{c|}{\textbf{Chameleon}} & \multicolumn{4}{c}{\textbf{GPT-4o}} \\
            \cmidrule(lr){4-7} \cmidrule(lr){8-11} \cmidrule(lr){12-15}
            &&& {\multirow{2.5}{*}{\textbf{Clean}}} & \multicolumn{3}{c|}{\textbf{Attacked}} & {\multirow{2.5}{*}{\textbf{Clean}}} & \multicolumn{3}{c|}{\textbf{Attacked}} & {\multirow{2.5}{*}{\textbf{Clean}}} & \multicolumn{3}{c}{\textbf{Attacked}} \\
            \cmidrule(lr){5-7} \cmidrule(lr){9-11} \cmidrule(lr){13-15}
            &&&& \textbf{No Deg.} & \textbf{Med Deg.} & \textbf{High Deg.} && \textbf{No Deg.} & \textbf{Med Deg.} & \textbf{High Deg.} && \textbf{No Deg.} & \textbf{Med Deg.} & \textbf{High Deg.} \\
            \midrule
            \multirow{13.7}{*}{PGD} 
            & \multirow{3}{*}{UnivFD} 
            & DRCT-CLIP & 95.9 & 15.9 & 57.2 & 62.2 & 79.2 & 21.8 & 52.2 & 63.9 & 99.2 & 16.4 & 75.2 & 66.0 \\
            && DRCT-ConvB & 95.9 & 76.8 & 81.9 & 67.1 & 79.2 & 81.4 & 79.3 & 68.7 & 99.2 & 83.4 & 94.1 & 69.1 \\
            && Corvi & 95.9 & 76.2 & 80.3 & 71.3 & 79.2 & 73.7 & 72.6 & 68.8 & 99.2 & 82.6 & 92.5 & 70.7 \\
            \cmidrule{2-15}
            & \multirow{3}{*}{DRCT-CLIP} 
            & UnivFD & 97.3 & 8.7 & 45.4 & 60.8 & 70.7 & 8.8 & 43.6 & 70.4 & 95.0 & 5.0 & 26.5 & 46.0 \\
            && DRCT-ConvB & 97.3 & 64.6 & 73.3 & 62.2 & 70.7 & 70.2 & 77.3 & 75.8 & 95.0 & 52.0 & 55.0 & 49.4 \\
            && Corvi & 97.3 & 72.8 & 78.1 & 64.9 & 70.7 & 67.8 & 73.9 & 74.7 & 95.0 & 50.3 & 63.0 & 50.5 \\
            \cmidrule{2-15}
            & \multirow{3}{*}{DRCT-ConvB} 
            & UnivFD & 98.8 & 75.6 & 61.0 & 53.2 & 69.6 & 33.3 & 32.5 & 41.1 & 88.5 & 66.6 & 67.3 & 46.6 \\
            && DRCT-CLIP & 98.8 & 24.8 & 50.2 & 55.2 & 69.6 & 5.6 & 28.2 & 40.2 & 88.5 & 18.8 & 56.9 & 45.3 \\
            && Corvi & 98.8 & 1.4 & 6.9 & 48.0 & 69.6 & 0.0 & 2.4 & 40.0 & 88.5 & 0.7 & 7.2 & 40.5 \\
            \cmidrule{2-15}
            & \multirow{3}{*}{Corvi} 
            & UnivFD & 100.0 & 95.5 & 82.4 & 57.2 & 63.0 & 27.3 & 25.3 & 33.2 & 93.3 & 67.1 & 54.3 & 57.3 \\
            && DRCT-CLIP & 100.0 & 94.0 & 83.9 & 58.0 & 63.0 & 26.2 & 29.9 & 37.4 & 93.3 & 62.5 & 59.9 & 57.3 \\
            && DRCT-ConvB & 100.0 & 96.3 & 86.6 & 62.6 & 63.0 & 49.8 & 42.3 & 38.5 & 93.3 & 83.0 & 73.7 & 59.6 \\
            \cmidrule{2-15}
            & Average & {\textendash} & 98.0 & 58.6 & 65.6 & 60.2 & 70.6 & 38.8 & 46.6 & 54.4 & 94.0 & 49.0 & 60.5 & 54.9 \\
            \midrule
            \multirow{4}{*}{Ensemble} 
            & UnivFD & LTO & 95.9 & 0.0 & 45.6 & 64.7 & 79.2 & 0.0 & 43.4 & 61.8 & 99.2 & 0.0 & 65.6 & 64.5 \\
            & DRCT-CLIP & LTO & 97.3 & 0.9 & 50.5 & 62.6 & 70.7 & 0.9 & 54.6 & 72.0 & 95.0 & 0.1 & 31.5 & 46.7 \\
            & DRCT-ConvB & LTO & 98.8 & 8.1 & 26.8 & 50.2 & 69.6 & 0.9 & 6.8 & 36.4 & 88.5 & 6.3 & 29.5 & 44.8 \\
            & Corvi & LTO & 100.0 & 84.6 & 76.3 & 54.1 & 63.0 & 19.3 & 28.2 & 34.9 & 93.3 & 47.0 & 53.0 & 54.8 \\
            \cmidrule{2-15}
            & Average & {\textendash} & 98.0 & 23.4 & 49.8 & 57.9 & 70.6 & 5.3 & 33.3 & 51.3 & 94.0 & 13.4 & 44.9 & 52.7 \\
            \midrule
            \multirow{4}{*}{Diverse} 
            & UnivFD & LTO & 95.9 & 0.0 & 6.4 & 51.7 & 79.2 & 0.1 & 3.7 & 43.9 & 99.2 & 0.1 & 8.0 & 48.2 \\
            & DRCT-CLIP & LTO & 97.3 & 0.6 & 7.6 & 52.9 & 70.7 & 0.6 & 9.3 & 59.6 & 95.0 & 0.1 & 4.0 & 37.3 \\
            & DRCT-ConvB & LTO & 98.8 & 1.5 & 7.6 & 43.0 & 69.6 & 0.1 & 2.0 & 28.3 & 88.5 & 2.7 & 12.8 & 39.7 \\
            & Corvi & LTO & 100.0 & 14.4 & 23.4 & 43.7 & 63.0 & 0.4 & 4.8 & 24.8 & 93.3 & 3.1 & 14.8 & 49.1 \\
            \cmidrule{2-15}
            & Average & {\textendash} & 98.0 & 4.1 & 11.3 & 47.8 & 70.6 & 0.3 & 5.0 & 39.2 & 94.0 & 1.5 & 9.9 & 43.6 \\
            \midrule
            \multirow{4}{*}{Universal} 
            & UnivFD & LTO & 95.9 & 20.8 & 74.0 & 63.6 & 79.2 & 31.5 & 76.8 & 69.3 & 99.2 & 33.7 & 87.2 & 62.9 \\
            & DRCT-CLIP & LTO & 97.3 & 56.9 & 64.8 & 61.6 & 70.7 & 49.9 & 74.6 & 74.3 & 95.0 & 39.0 & 57.4 & 47.6 \\
            & DRCT-ConvB & LTO & 98.8 & 23.9 & 24.4 & 51.4 & 69.6 & 5.8 & 18.3 & 38.6 & 88.5 & 9.3 & 26.6 & 46.4 \\
            & Corvi & LTO & 100.0 & 85.8 & 65.2 & 53.4 & 63.0 & 22.3 & 34.6 & 37.6 & 93.3 & 62.0 & 60.5 & 54.7 \\
            \cmidrule{2-15}
            & Average & {\textendash} & 98.0 & 46.9 & 57.1 & 57.5 & 70.6 & 27.4 & 51.1 & 55.0 & 94.0 & 36.0 & 57.9 & 52.9 \\
            \midrule
            \multirow{4}{*}{Query} 
            & UnivFD & {\textendash} & 95.9 & 64.1 & 74.7 & 66.9 & 79.2 & 60.4 & 64.2 & 69.0 & 99.2 & 74.1 & 91.0 & 70.3 \\
            & DRCT-CLIP & {\textendash} & 97.3 & 79.0 & 75.9 & 64.1 & 70.7 & 61.0 & 72.1 & 75.7 & 95.0 & 72.9 & 60.5 & 48.1 \\
            & DRCT-ConvB & {\textendash} & 98.8 & 77.2 & 63.1 & 54.8 & 69.6 & 33.5 & 35.2 & 42.3 & 88.5 & 63.4 & 64.8 & 47.9 \\
            & Corvi & {\textendash} & 100.0 & 98.5 & 86.0 & 51.9 & 63.0 & 24.6 & 31.4 & 37.2 & 93.3 & 70.4 & 55.0 & 57.0 \\
            \cmidrule{2-15}
            & Average & {\textendash} & 98.0 & 79.7 & 74.9 & 59.4 & 70.6 & 44.9 & 50.7 & 56.1 & 94.0 & 70.2 & 67.8 & 55.8 \\
            \bottomrule
        \end{tabular}
        }    
        \label{tab:attack_results}
    \end{table*}

\begin{table}[t]
\centering
\caption{PGD step convergence analysis. Post-attack AUROC (\%) averaged over three target detectors in black-box setting ($\epsilon=8/255$, $\alpha=2/255$). Lower values indicate more effective attacks. The negligible differences across step counts confirm that ten steps is sufficient for convergence.}
\label{tab:step_convergence}
\small
\begin{tabular}{lccc}
\toprule
\textbf{Source} & \textbf{5 steps} & \textbf{10 steps} & \textbf{20 steps} \\
\midrule
Corvi & 83.4 & 84.3 & 83.7 \\
DRCT-CLIP & 49.8 & 49.1 & 48.5 \\
DRCT-ConvB & 91.2 & 91.6 & 91.8 \\
UnivFD & 99.7 & 99.2 & 99.1 \\
\midrule
\textbf{Average} & \textbf{81.0} & \textbf{81.0} & \textbf{80.8} \\
\bottomrule
\end{tabular}
\end{table}
    
\begin{figure*}[h]
    \centering
    \includegraphics[width=.98\textwidth]{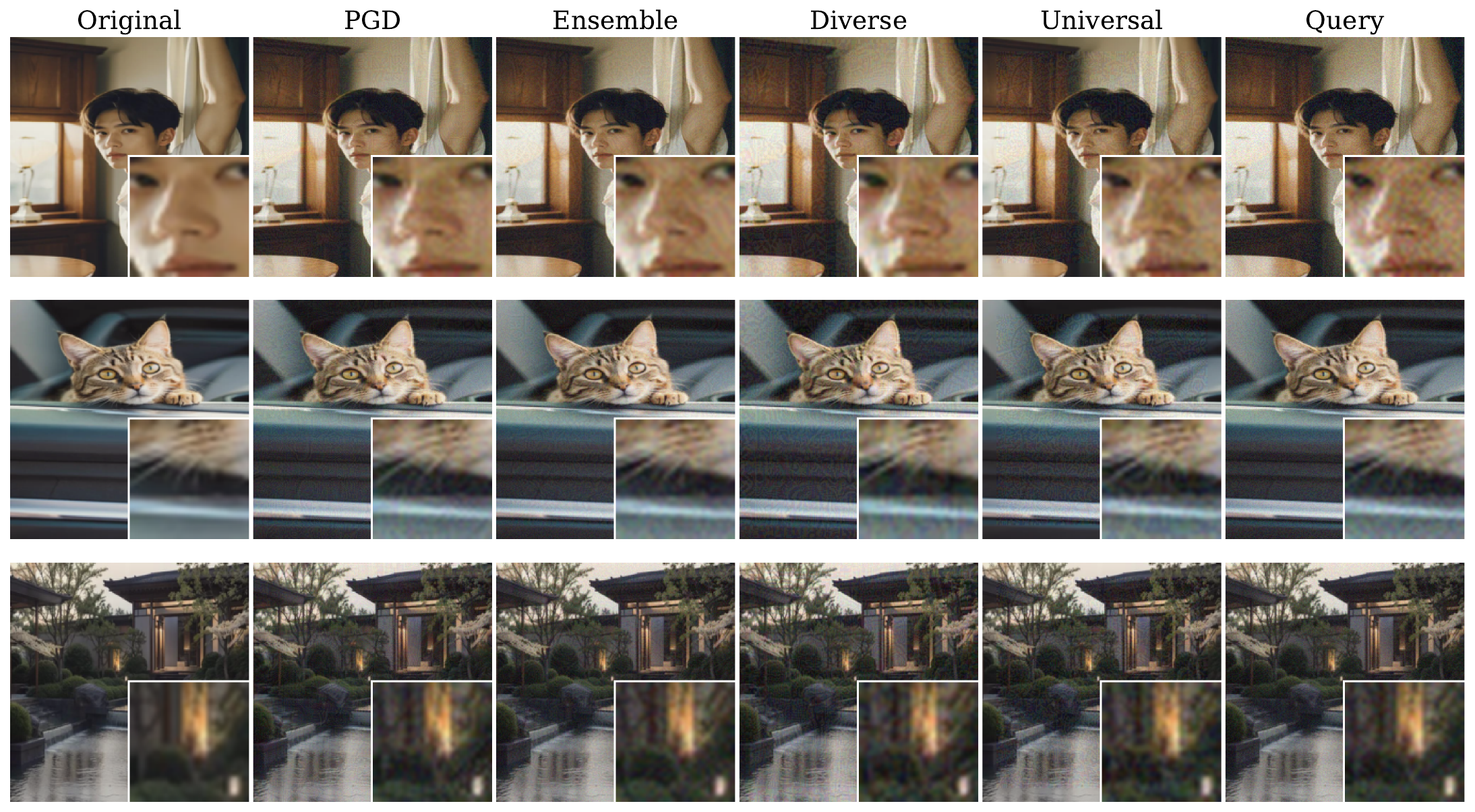}
    \caption{Adversarial examples generated for target UnivFD with $\epsilon=8/255$. Images are from the Chameleon dataset. }
    \label{fig:qual}
\end{figure*}

\end{document}